\documentclass[onecolumn]{robbyant}           %

\usepackage{makecell}
\usepackage{wrapfig}
\usepackage{subcaption}
\usepackage{tikz}
\usepackage{pgf-pie}
\usepackage{pgfplots}
\pgfplotsset{compat=1.18}
\definecolor{pie1}{HTML}{5B8FF9}
\definecolor{pie2}{HTML}{5AD8A6}
\definecolor{pie3}{HTML}{F6BD16}
\definecolor{pie4}{HTML}{E86452}
\definecolor{pie5}{HTML}{6DC8EC}
\definecolor{pie6}{HTML}{945FB9}
\definecolor{pie7}{HTML}{FF9845}
\definecolor{pie8}{HTML}{1E9493}
\definecolor{pie9}{HTML}{FF99C3}

\metadata[Website]{\url{https://technology.robbyant.com/lingbot-depth}}
\metadata[Github]{\url{https://github.com/robbyant/lingbot-depth}}
\metadata[Checkpoints]{\url{https://huggingface.co/robbyant/lingbot-depth}}
\newcommand{\methodname}{LingBot-Depth}
\newcommand{\method}{\texttt{\methodname}\xspace}
\newcommand{\methods}{\texttt{\methodname-S}\xspace}
\newcommand{\methodr}{\texttt{\methodname-R}\xspace}

\title{Masked Depth Modeling for Spatial Perception}

\author{
\begin{center}
    Bin Tan \quad
    Changjiang Sun \quad
    Xiage Qin \quad
    Hanat Adai \quad
    Zelin Fu \quad
    Tianxiang Zhou
    \\[5pt]
    Han Zhang \quad
    Yinghao Xu \quad
    Xing Zhu \quad
    Yujun Shen \quad
    Nan Xue$^\dagger$
    \\[12pt]
    $^{\dagger}$Project Lead
\end{center}
}

\begin{document}

\abstract{%
Spatial visual perception is a fundamental requirement in physical-world applications like autonomous driving and robotic manipulation, driven by the need to interact with 3D environments.
Capturing pixel-aligned metric depth using RGB-D cameras would be the most viable way, yet it usually faces obstacles posed by hardware limitations and challenging imaging conditions, especially in the presence of specular or texture-less surfaces.
In this work, we argue that the inaccuracies from depth sensors can be viewed as ``masked'' signals that inherently reflect underlying geometric ambiguities.
Building on this motivation, we present \method, a depth completion model which leverages visual context to refine depth maps through \textit{masked depth modeling} and incorporates an automated data curation pipeline for scalable training.
It is encouraging to see that our model outperforms \textbf{\textit{top-tier RGB-D cameras}} in terms of both depth precision and pixel coverage.
Experimental results on a range of downstream tasks further suggest that \method offers an aligned latent representation across RGB and depth modalities.
We release the code, checkpoint, and $3M$ RGB-depth pairs (including $2M$ real data and $1M$ simulated data) to the community of spatial perception.
}

\maketitle

\justifying
\section{Introduction}\label{sec:intro}
Precise perception of the 3D world is fundamental to any intelligent agent operating in the physical environment—from biological organisms to autonomous vehicles and generalist robots. It provides the essential context for localization (``where am I?'') and scene understanding (``what is around me?'').
Without robust 3D perception, real-world actions cannot be reliably planned, executed, or verified. Consequently, the pursuit of accurate 3D sensing has become a central pillar of research on physical-grounded artificial intelligence.
The optimal approach to achieving this goal remains a subject of debate, with current methods generally falling into three paradigms: (1) classic multi-view visual geometry and recent learning-based adventures, (2) data-driven monocular depth estimation, and (3) active sensor-based depth measurement (\textit{e.g.}, LiDAR, ToF, Structured Light). 
While the trade-offs of each paradigm have been extensively discussed~\cite{FoundationStereo,promptda,moge2}, the requirements for effective perception came down to three critical criteria: (1) absolute metric scale, (2) pixel-aligned dense geometry, and (3) real-time acquisition without computationally expensive post-processing.

\begin{figure}[t]
    \centering
    \includegraphics[width=\linewidth]{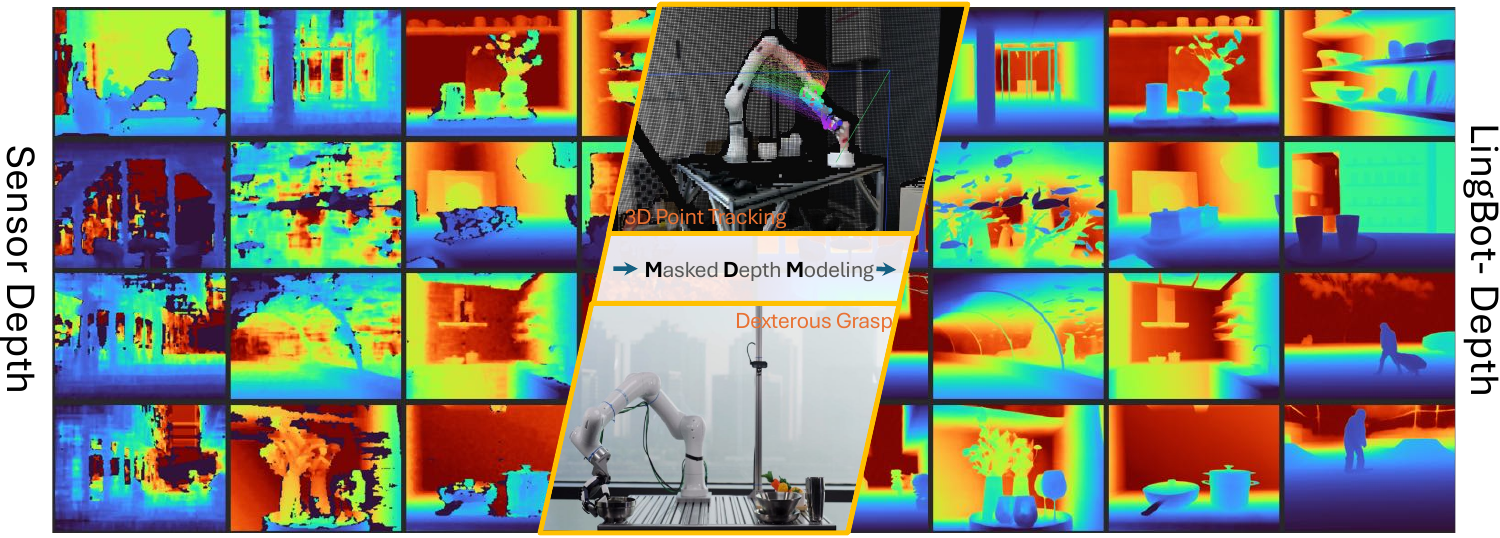}
    \vspace{-4mm}
    \caption{{\bf Enhanced sensor depth} powered by our proposed MDM pretraining, which leverages naturally missing depth measurements in RGB-D sensors as masks to learn metric-scale, complete, and accurate depth representations. The resulting \method model serves as a powerful spatial perception prior for downstream applications, including 3D point tracking and dexterous grasping.}
    \vspace{-2mm}
    \label{fig:teaser}
\end{figure}

RGB-D cameras distinguish themselves as the only modality capable of satisfying these requirements in real-time. However, their utility is frequently compromised by inherent hardware limitations, particularly the susceptibility of stereo matching algorithms to appearance ambiguities.
As illustrated in the left of \cref{fig:teaser}, even state-of-the-art commercial sensors struggle in challenging scenarios—such as low-texture surfaces, specular reflections, and complex lighting conditions. These failures manifest as severe data corruption and missing values, directly violating the requirement for dense, pixel-aligned geometry.

In this work, we propose a paradigm shift: rather than treating these sensor failures as noise to be discarded, we leverage them as a useful learning signal.
Inspired by recent advances in self-supervised masked modeling~\cite{MAE} and joint embedding architectures~\cite{jepa}, we introduce {\bf M}asked {\bf D}epth {\bf M}odeling (MDM) to achieve dense, pixel-aligned scene geometry. A key innovation of MDM is the interpretation of missing regions (``holes'') in raw depth maps as ``natural masks'', diverging from standard MAE approaches that rely solely on random masking.
Because these natural masks arise specifically from geometric and appearance ambiguities (\textit{e.g.}, specular reflections), they present a significantly harder reconstruction challenge than random dropout. To solve this, our architecture provides the full, unmasked RGB image as a condition. The model is thus forced to infer the missing depth values by reasoning about the correlation between the complete RGB context and the remaining valid depth tokens.

By unifying the objectives of monocular depth estimation and depth completion, our MDM framework serves as a versatile generalist capable of yielding metric-scale, pixel-aligned, dense depth maps from arbitrary RGB-D inputs.
The transition between these tasks is governed simply by the masking strategy within the Transformer. In the extreme case where all depth tokens are masked, the model functions as a pure Monocular Depth Estimator, forcing the Self-Attention layers to exploit RGB context alone to infer geometry. Conversely, for Depth Completion, we mask only the invalid (sensor-corrupted) tokens, allowing the model to fuse the sparse valid depth readings with visual cues to reconstruct a complete, dense depth prediction.

To support large-scale MDM training, we built a scalable data curation pipeline that bridges raw sensor data and reliable supervision. The pipeline includes two parallel streams: a synthetic branch based on self-hosted 3D assets, and a real-world branch using a modular, 3D-printed capture rig compatible with a variety of consumer RGB-D cameras, including active stereo (Intel RealSense, Orbbec Gemini) and passive stereo (ZED) systems.
Using this setup, we collected 1M synthetic samples and 2M real captures, each containing synchronized RGB images, raw sensor depth, and stereo pairs. The stereo pairs enable pseudo-depth supervision through a custom stereo matching network adapted from FoundationStereo~\cite{FoundationStereo} and trained on synthetic data. We further enrich this corpus with several public RGB-D datasets~\cite{tartanair,adt,arkitscenes,scannet++,hypersim,clear_grasp,taskonomy}, forming a diverse training set for robust depth completion.
Pretraining a ViT-Large on this curated RGB-D corpus with Masked Depth Modeling allows metric geometry to be integrated into semantic tokens via attention, thus improving the sensing quality of RGB-D cameras, as illustrated in the right panel of \cref{fig:teaser}. 

Experimentally, we first validate MDM on standard benchmarks, where it achieves competitive performance in both depth completion and metric monocular depth estimation. Moreover, it serves as a stronger monocular depth prior for FoundationStereo~\cite{FoundationStereo} compared to DepthAnythingV2~\cite{DepthAnythingV2}.
Despite being trained solely on static images, our model demonstrates remarkable zero-shot generalization to video depth estimation, producing temporally consistent geometry without explicit temporal supervision. Leveraging this spatiotemporal consistency, we deploy our pretrained model as a drop-in depth estimator for 3D tracking, replacing the standard VGGT~\cite{vggt} frontend in SpatialTrackerV2~\cite{spatracker2}. This leads to improved motion understanding and higher computational efficiency in real-world scenarios.
Finally, we demonstrate practical robotic utility by training a dexterous grasping policy. Unlike prior approaches that depend on object CAD models or perfect simulator states, our policy leverages the robust depth predictions and latent representations learned through MDM to enable open-world grasping. We successfully grasp challenging objects that typically defeat conventional sensors, including transparent glassware and highly reflective bowls.

\section{Masked Depth Modeling}\label{sec:method}

\begin{figure}[t]
    \centering
    \includegraphics[width=0.9\linewidth]{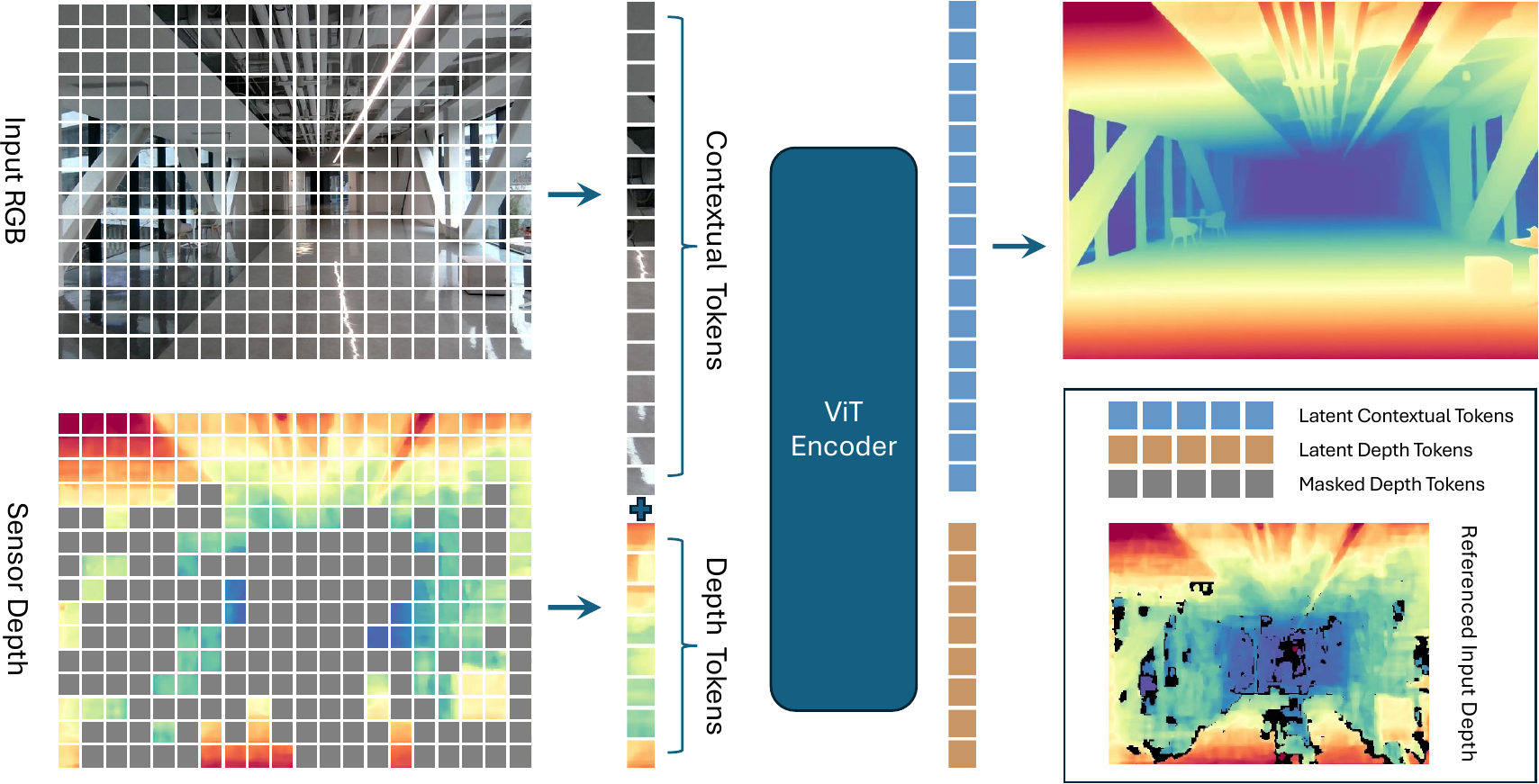}
    \vspace{-2mm}
    \caption{
    {\bf Illustration of the proposed Masked Depth Modeling framework}. 
    Depth tokens corresponding to missing sensor measurements are masked, and a ViT encoder learns a joint embedding from the contextual tokens (i.e., the RGB frame) and the remaining unmasked depth tokens.
    In the decoder stage, latent depth tokens are discarded, and a ConvStack decoder reconstructs the full depth map from latent contextual tokens. 
    We put an unmasked depth map in the bottom-right as the reference. 
    }
    \vspace{-3mm}
    \label{fig:mdm-method}
\end{figure}

Masked depth modeling follows the general paradigm of masked image modeling \cite{MAE} within an encoder–decoder framework, but shifts the learning objective from appearance reconstruction to depth map prediction by operating on RGB-D inputs.
We adopt a standard Vision Transformer (ViT) architecture \cite{vit}, using ViT-Large as the encoder backbone.
For depth reconstruction from patch tokens, instead of the shallow Transformer decoder used in vanilla MAE \cite{MAE}, we employ a ConvStack decoder adopted from MoGe \cite{moge,moge2}, which is better suited for dense geometric prediction.
\Cref{fig:mdm-method} provides an overview of the masked depth modeling architecture.

\subsection{Separated Patch Embedding for RGB-D Inputs}
\paragraph{Patch Embedding Layers.} 
We apply separate patch embedding layers to the two input modalities: the 3-channel RGB image and the single-channel depth map. Each modality is independently projected into a sequence of patch tokens, yielding RGB tokens and depth tokens that are spatially aligned on the same 2D grid.
This separated patch embedding design enables the self-attention layers in the ViT encoder to learn joint representations that integrate appearance context from RGB images with geometric cues from depth measurements. In particular, the model can exploit complementary information from rich visual context and fundamental geometric priors such as near--far relationships, coplanarity, and spatial continuity.
In our implementation, we set the patch size to 14 for both patch embedding layers, following DINOv2~\cite{DINOv2}. Without loss of generality, we assume that the RGB frame and the depth map share the same spatial resolution $(H, W)$, where both $H$ and $W$ are divisible by 14. The number of tokens per modality is therefore $N = HW / 14^2$. We denote the $i$-th RGB token as $\mathbf{c}_i \in \mathbb{R}^n$ and the $i$-th depth token as $\mathbf{d}_i \in \mathbb{R}^n$, where $n$ is the token embedding dimension.

\paragraph{Positional Embeddings.} 
Compared to standard ViT inputs, RGB-D token sequences require encoding two types of positional information:
(1) the 2D spatial location of each token, and (2) the modality identity, distinguishing RGB tokens from depth tokens at the same spatial position.
To this end, we introduce two types of positional embeddings:
(1) a shared learnable 2D spatial positional embedding for both RGB and depth tokens, capturing their spatial locations in the image plane; and (2) a modality embedding that distinguishes the input source of each token—RGB or depth. Specifically, the modality embedding is set to 1 for RGB tokens and 2 for depth tokens.
The final positional encoding of each token is computed as the sum of its spatial and modality embeddings. 
Finally, each RGB/depth token $\mathbf{c}_i$ or $\mathbf{d}_i$ are added with the corresponding positional embedding before feeding to the Attention blocks.

\subsection{Joint Embedding of RGB and Unmasked Depth for Masked Depth Prediction}
Depth sensors are inherently sensitive to appearance-related disturbances, and it is common for RGB-D cameras to produce depth maps with missing or invalid measurements.
Importantly, such missing depth values often correlate with challenging visual conditions in the scene, including surface material properties, lighting variations, and reflective or textureless regions.
Rather than treating these missing measurements as noise, we view them as an opportunity to learn a joint representation from the complementary sources: valid depth observations and the always-available RGB appearance context.
Accordingly, we leverage the masking patterns naturally induced by depth sensors and train the model to learn a joint embedding of RGB tokens and unmasked depth tokens, enabling robust depth reasoning under incomplete observations.

\paragraph{Masking from Missing Depth Measurements.}
Our masking strategy is grounded in the inductive bias introduced by missing depth measurements. In practice, however, many RGB-D samples are well conditioned and contain few or no missing depth values. Such samples are nonetheless valuable for learning joint embeddings, as they provide abundant paired appearance--geometry observations and should not be discarded. Accordingly, we aim to incorporate as many RGB-D samples as possible to support large-scale training of Vision Transformers.
Another practical consideration arises from patch-based processing: a single patch may contain a mixture of valid and invalid depth values. To address this ambiguity, we define the masking decision at the patch level based on the validity statistics within each patch, ensuring a consistent and well-defined masking rule for depth tokens.

Based on the above discussion, a depth patch (token) whose values are entirely missing is always masked. For patches containing a mixture of valid and invalid depth values, we assign a higher probability ({\em e.g.}, 0.75 in our training) of being masked. If the number of masked tokens from these two cases is insufficient to meet the target masking ratio, we randomly sample additional fully valid depth tokens to complete the masking set. Such a masking strategy allows imperfect yet informative depth tokens to remain unmasked, enabling meaningful interaction with contextual RGB tokens. 
The overall masking ratio is in the range of 60\%-90\% for depth maps. 

\paragraph{RGB-D Tokens for Vision Transformers.}
After applying the masking strategy, the full set of RGB tokens and the unmasked depth tokens are concatenated to form RGB-D tokens, which are fed into a ViT-Large encoder with 24 self-attention blocks. In addition to the RGB-D tokens, a \texttt{[cls]} token is retained to capture global context across modalities.
Different from conventional designs that aggregate tokens from multiple intermediate layers (e.g., layers 6, 12, 18, and 24) for dense prediction, as adopted in DepthAnythingV2~\cite{DepthAnythingV2} and MoGe~\cite{moge,moge2}, we only retain the output tokens from the final encoder layer for subsequent processing.

In vanilla MAE, pixel-wise RGB reconstruction is performed using a shallow Transformer decoder. Each patch token is decoded into $3 \times \texttt{patch\_size} \times \texttt{patch\_size}$ values, which are then reshaped into the pixel domain. This design is suitable for RGB-only masking, where masked tokens share homogeneous latent representations within the decoder.
In contrast, our masking strategy is applied exclusively to depth tokens, while the RGB tokens remain fully observable and provide complete contextual information. This setting allows depth prediction at each spatial location to be conditioned on rich appearance context. Consequently, we adopt a convolutional decoder architecture, termed \emph{ConvStack}, which is more appropriate for dense geometric reconstruction.
\begin{figure}[t]
    \centering
    \includegraphics[width=0.95\linewidth]{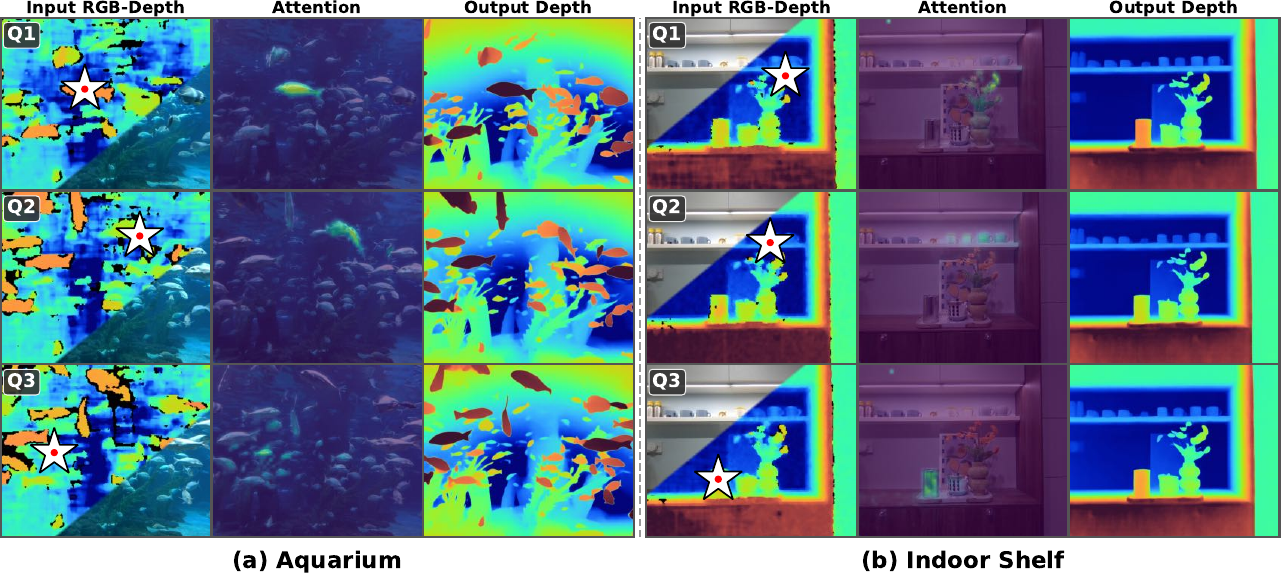}
    \caption{
    {\bf Multi-query depth-to-RGB attention visualization.} 
    For two scenes, (a) an aquarium with densely packed objects and (b) an indoor shelf with heterogeneous materials, we select three depth query patches (Q1--Q3) and visualize their attention over RGB tokens. Each row shows the masked input depth (with query location marked by $\star$), the attention overlay on the RGB image, and the refined depth output. Different queries attend to distinct, spatially corresponding regions, confirming that the joint embedding captures fine-grained cross-modal geometric--appearance associations. The RGB-D camera we used here is Orbbec Gemini-335.
    }
    \label{fig:attention-vis}
\end{figure}

\paragraph{ConvStack Decoder.}
After the encoder, latent depth tokens are discarded, while the latent contextual tokens are retained as spatially distributed representations. To inject global scene context, the \texttt{[cls]} token is broadcast and element-wise added to each contextual token, enriching them with task-level semantic information. These enhanced tokens serve as input queries to a hierarchical convolutional decoder adapted from MoGe~\cite{moge,moge2}.

The decoder follows a pyramid structure composed of a shared convolutional neck and multiple task-specific heads. Starting from a low-resolution feature map of size $(h, w)$, the neck progressively upsamples features through stacked residual blocks and transposed convolutions (kernel size 2, stride 2), doubling the spatial resolution at each stage until reaching $(16h, 16w)$. At each scale, UV positional encodings—derived from a circular mapping of image coordinates—are injected to preserve spatial layout and aspect ratio.
The resulting multi-scale feature pyramid is shared across all task heads, enabling efficient feature reuse while allowing each head to decode its own dense output. The final depth prediction is bilinearly upsampled to match the original input resolution.
This design decouples high-level context modeling from dense geometric reconstruction, enabling precise depth estimation while supporting scalable and flexible multi-task learning.

\paragraph{Attention Visualization.}
To verify that the joint embedding effectively captures cross-modal associations, we visualize the depth-to-RGB attention maps from the final encoder layer. For a selected depth token, we extract its attention weights over all RGB tokens and project them back onto the image plane as a heatmap overlay.
As shown in \cref{fig:attention-vis}, we select multiple query patches within the same scene and visualize their respective attention patterns. Across diverse scenarios—an aquarium with densely packed objects at varying depths, and an indoor shelf with heterogeneous materials—different depth tokens consistently attend to distinct, spatially localized regions in the RGB image that correspond to their respective query locations. This demonstrates that the encoder learns fine-grained, position-aware geometric--appearance correspondences through the masked depth modeling objective, rather than collapsing to global or trivial attention patterns.

\subsection{Training Details}
We adopt a 24-layer ViT-Large encoder (ViT-L/14)~\cite{DINOv2} as the visual backbone.
Since training Vision Transformers is data-intensive and our goal is to explore joint contextual modeling of RGB frames and depth maps, we initialize the encoder with the official DINOv2 pretrained checkpoint.
The decoder, which consists of a shared convolutional neck and task-specific heads, is randomly initialized.
To accommodate the different optimization dynamics of the pretrained encoder and the randomly initialized decoder, we employ a differential learning rate strategy: parameters in the encoder backbone are optimized with a base learning rate of $1 \times 10^{-5}$, while all remaining parameters, including those in the decoder, are trained with a higher learning rate of $1 \times 10^{-4}$.
The optimizer is AdamW~\cite{loshchilov2017decoupled} with momentum parameters $\beta_1 = 0.9$, $\beta_2 = 0.999$, and a weight decay of 0.05. Parameter groups are defined based on module names, such that all parameters matching the pattern \texttt{*backbone*} are assigned to the low-learning-rate group.
We employ a composite learning rate schedule. During the first 2{,}000 iterations, the encoder learning rate is linearly warmed up from zero to $1 \times 10^{-5}$, while the decoder learning rate starts directly at its target value. After warm-up, a step decay scheduler reduces both learning rates by a factor of 0.5 every 25{,}000 iterations.

Training is conducted for a total of 250{,}000 iterations with a global batch size of 1{,}024, achieved using 128 GPUs and a per-GPU batch size of 8. Data augmentation includes random resized cropping and horizontal flipping, along with a set of synthetic image degradations—specifically color jittering, JPEG compression artifacts, motion blur, and shot noise—to improve robustness under realistic visual conditions.
Gradient clipping with a maximum norm of 1.0 is applied to stabilize optimization, and mixed-precision training using BF16 is enabled throughout to improve computational efficiency and reduce memory consumption. The entire training process takes approximately 7.5 days.
The predicted depth map is supervised using an L1 loss applied directly to the ground-truth depth map. Only pixels with valid depth values in the ground truth are included in the loss computation.

\begin{figure}[t]
    \centering
    \includegraphics[width=0.95\linewidth]{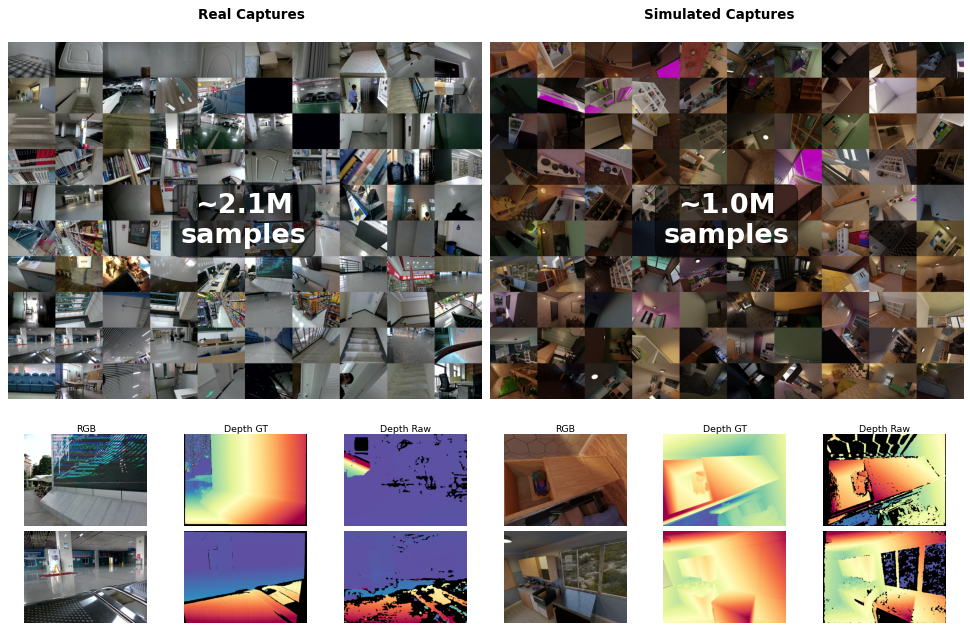}
    \caption{{\bf Our data curation pipelines.} Samples from a total of 2.1M real-captured samples plus 1.0M simulated captures are gather in the top row. In the bottom row, we show the RGB-D inputs and the GT depth maps accordingly.}
    \label{fig:MDM-dataset}
\end{figure}
\section{Data Curation Pipelines}\label{sec: data}
RGB-D data is considerably scarcer than RGB-only data due to the reliance on specialized depth sensors. 
\begin{wrapfigure}{r}{0.38\linewidth}
    \vspace{-10pt}
    \centering
    \includegraphics[width=0.85\linewidth]{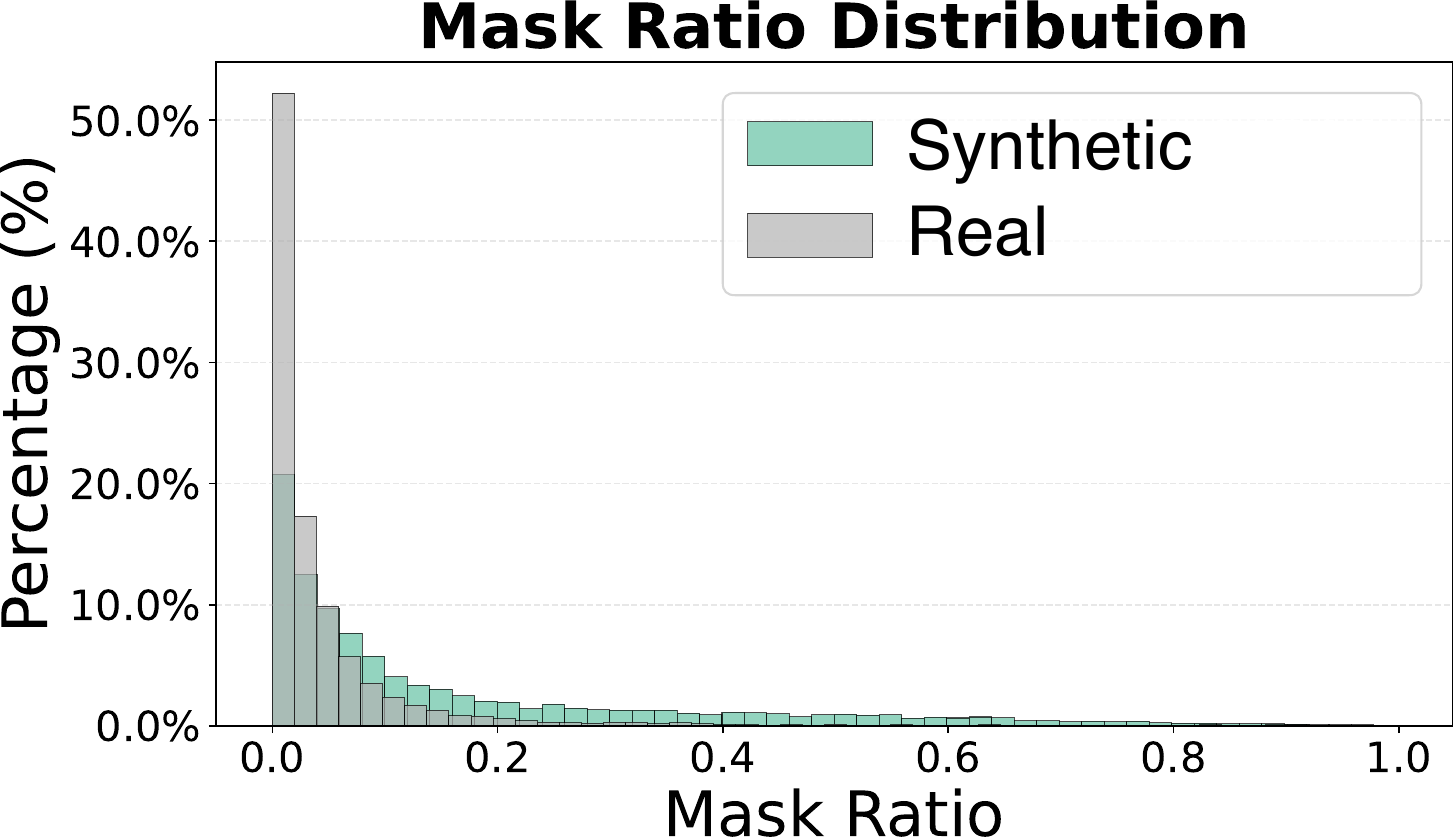}
    \vspace{-5pt}
    \caption{
    \textbf{Mask ratio distributions} for our curated synthetic and real-world RGB-D datasets.
    We compute the ratio of invalid depth pixels as the original mask ratio. Note that the synthetic data was processed by open-source SGM~\cite{sgm} algorithm, it has more missing measurements in the simulated sensor depth than the real captures.
    }
    \label{fig:histograms}
    \vspace{-5pt}
\end{wrapfigure}
Most existing RGB-D datasets either avoid challenging imaging conditions to reduce missing depth measurements or generate near-perfect depth maps using high-quality 3D assets and rendering engines. 
Consequently, they lack the naturally occurring depth incompleteness required for masked depth modeling. To overcome this limitation, we curate RGB-D data that preserves realistic missing patterns induced by real-world sensing. 
Our data curation pipeline consists of two parallel streams: a synthetic pipeline built on self-hosted 3D assets, and a real-world pipeline based on a scalable RGB-D capture system using multiple commercial depth cameras. The mask ratio distributions in the raw data of our data is shown in \cref{fig:histograms}. 
We name the synthetic data curated from our synthetic pipeline as \methods, and the real-world data curated from the real-world pipeline as \methodr.
In addition, we incorporate existing open-source RGB-D datasets as supplementary training data. For these datasets, we artificially corrupt depth maps with Gaussian noise and apply our masking strategy during training, while using the original depth maps as reconstruction targets with valid-pixel masks to exclude missing measurements.

\subsection{Synthetic Data Pipeline}
Our synthetic data pipeline differs from existing approaches that focus solely on rendering idealized depth maps. Instead, we explicitly simulate the imaging process of real-world active RGB-D cameras to generate realistic depth observations with natural imperfections.

Using self-hosted 3D assets, we simultaneously render RGB images, perfect depth maps, and grayscale stereo image pairs with speckle patterns in Blender. The RGB image is rendered from the left camera of the stereo pair, ensuring pixel-wise alignment between RGB appearance and the depth measurements produced by stereo matching. The rendered stereo images are then processed using the widely adopted semi-global matching (SGM) algorithm to produce sensor-like depth maps that mimic real-world capture artifacts.

For stereo rendering, we configure a pair of virtual cameras in Blender and randomly sample the stereo baseline from a uniform distribution between $0.05$ and $0.2$ meters. The camera focal length is independently sampled from a uniform distribution between $16$ and $28$ millimeters, enabling diverse imaging geometries.

\begin{figure}[t]
    \centering
    \includegraphics[width=.9\linewidth]{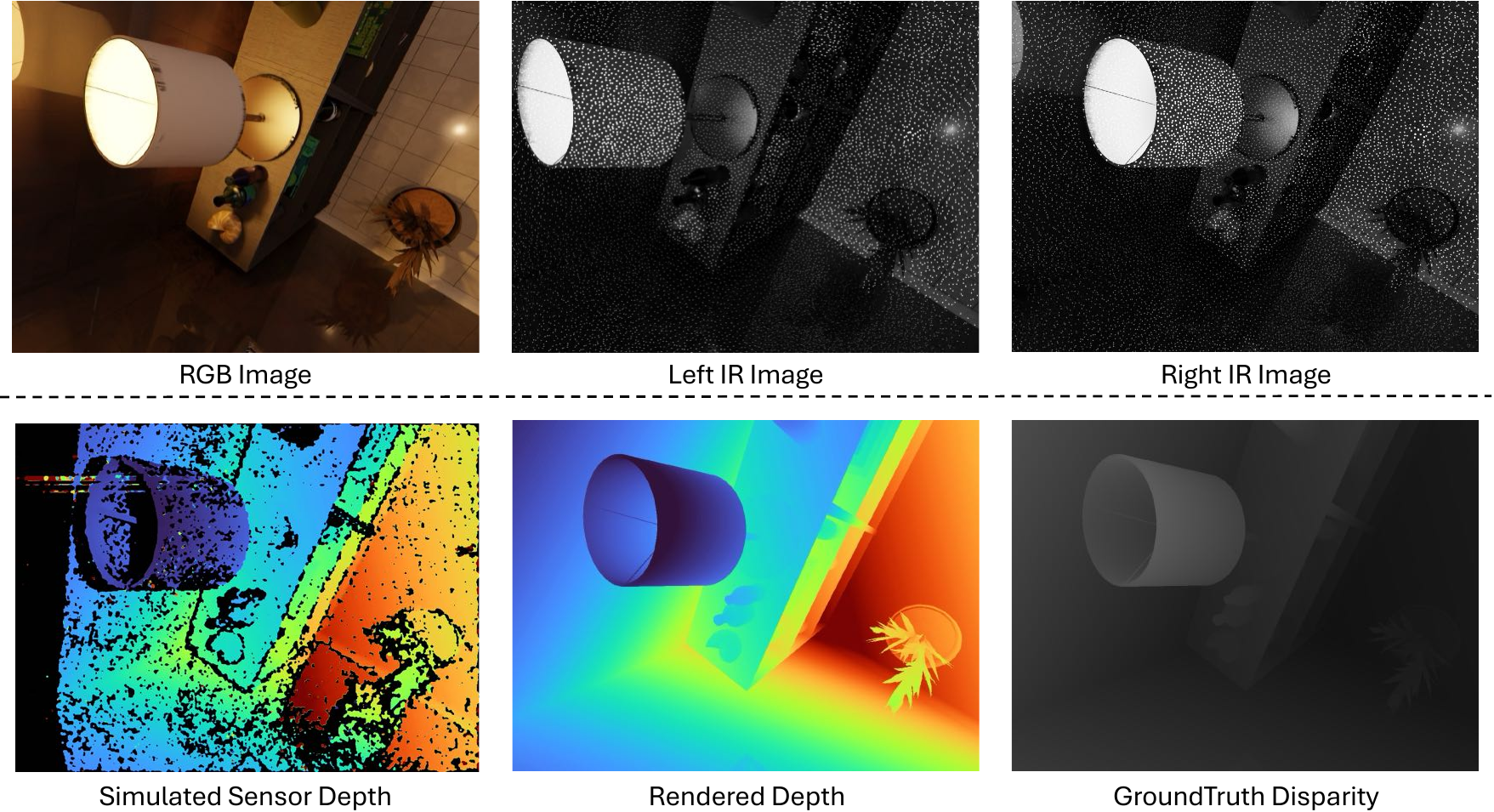}
    \caption{{\bf A synthetic data sample from our pipeline}. Each sample includes an RGB image, a perfect rendered depth, a stereo pair with speckle patterns, a ground-truth disparity map, and a simulated sensor depth computed via semi-global matching (SGM) to mimic real-world active depth camera artifacts.}
    \label{fig:sim-data}
\end{figure}
\paragraph{Resulting Data Sample Format.}
We render 10 million synthetic samples from 442 indoor scenes. Each sample (as shown in \cref{fig:sim-data}) contains the following components:
\begin{itemize}
    \item An \textbf{RGB image} rendered at a resolution of $960 \times 1280$ (H$\times$W);
    \item A \textbf{perfect depth map} rendered at the same target resolution;
    \item A \textbf{stereo image pair} rendered at a resolution of $720 \times 960$ and the groundtruth \textbf{disparity map} rendered at $960\times 1280$ resolution;
    \item A \textbf{sensor depth map} computed from the stereo pair at $720 \times 960$ and then upsampled using nearest-neighbor interpolation to match the target resolution.
\end{itemize}

\paragraph{Comparisons to Existing Datasets.}
Prior to our work, several efforts have explored the simulation of imperfect depth measurements~\cite{D3roma, Dreds}, but at a substantially smaller scale. For example, HSSD-IsaacSIM-STD~\cite{D3roma} renders approximately 10k stereo infrared image pairs, while DREDS~\cite{Dreds} generates about 130k stereo pairs. In contrast, our simulation pipeline produces data at a scale more than an order of magnitude larger.
Beyond data quantity, our dataset also differs qualitatively in terms of scene fidelity. We employ self-hosted, high-quality 3D scenes, rather than constructing scenes by simply stacking isolated objects in simulators, as done in DREDS~\cite{Dreds}. Although HSSD-IsaacSIM-STD~\cite{D3roma} adopts more realistic scene-level layouts, its reliance on robotics-oriented simulators introduces approximations that lead to reduced visual fidelity compared to real-world captures.

\subsection{Scalable RGB-D Capture Systems}
\begin{figure}[t]
    \centering
    \includegraphics[width=0.45\linewidth]{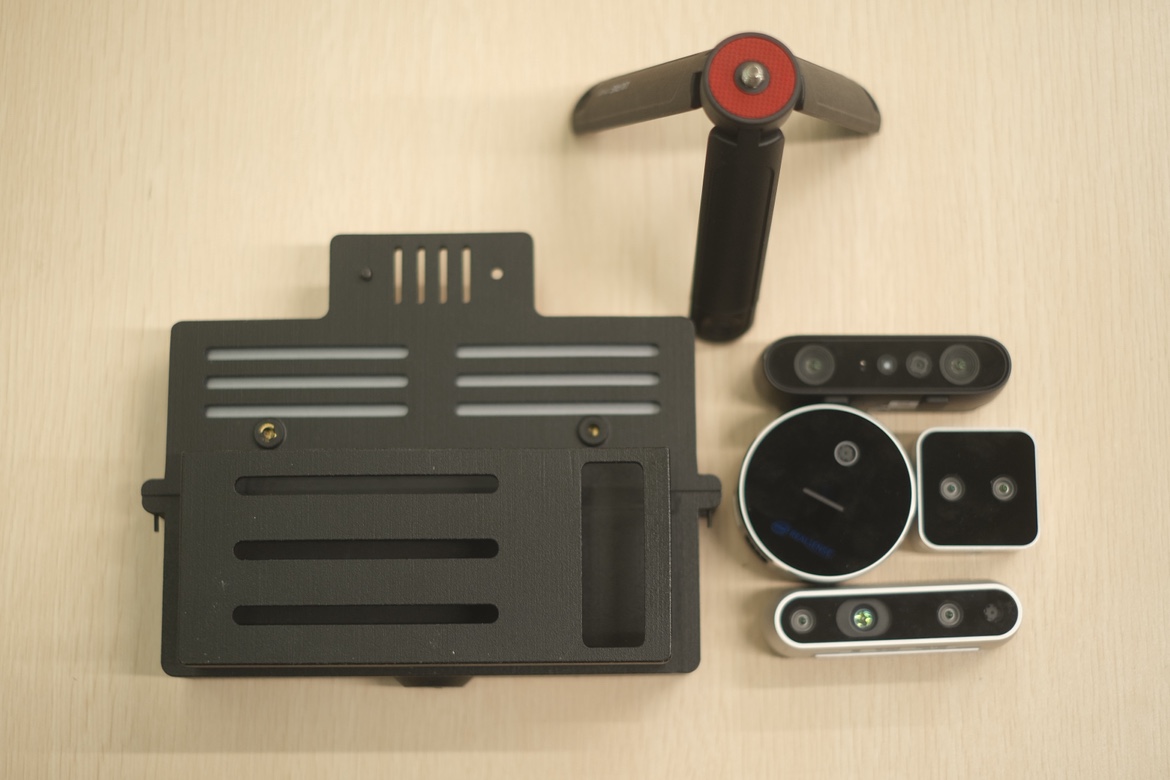}
    \hspace{5pt}
    \includegraphics[width=0.45\linewidth]{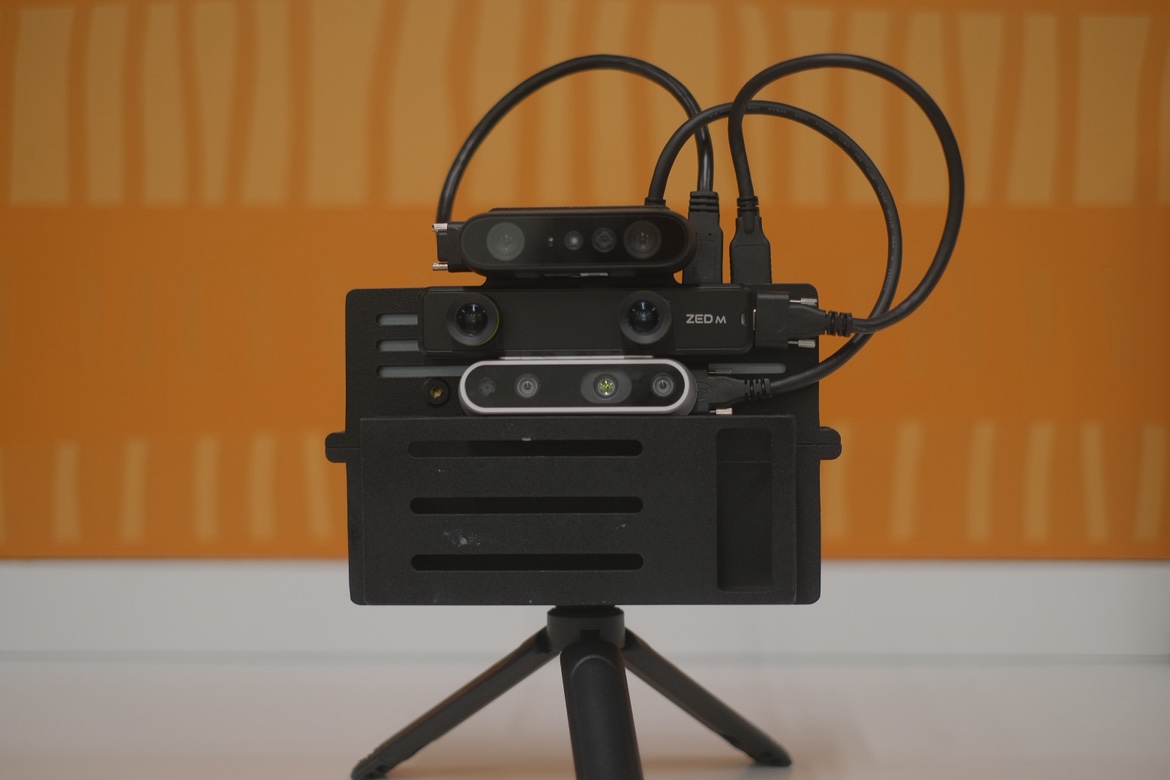}
    \vspace{-5pt}
    \caption{{\bf Our scalable RGB-D capture system}. Left: individual hardware components, including multiple RGB-D sensors, mounting fixtures, and supporting accessories. Right: the assembled capture system mounted on a tripod, enabling scalable and flexible real-world RGB-D data acquisition.}
    \label{fig:hardware-prototye}
    \vspace{-5pt}
\end{figure}

\begin{table}[t]
\centering
\caption{\textbf{Distribution of our real-world RGB-D captures by scene category.} We design a diverse collection protocol spanning residential, commercial, public, and specialized spaces to ensure broad coverage of indoor environments.}
\vspace{-5pt}
\label{tab:real_scene_distribution}

\small
\begin{tabular}{lr@{\hskip 30pt}lr@{\hskip 30pt}lr}
\toprule
\multicolumn{2}{c}{\textbf{Residential Space}} & \multicolumn{2}{c}{\textbf{Work / Study Space}} & \multicolumn{2}{c}{\textbf{Commercial / Service}} \\
\midrule
$< 50\,\mathrm{m}^2$           & 10.16\% & Office          & 3.39\% & Retail Store        & 3.39\% \\
$50\text{--}100\,\mathrm{m}^2$ & 10.16\% & Meeting Room    & 3.39\% & Restaurant / Caf\'{e} & 3.39\% \\
$> 100\,\mathrm{m}^2$          & 10.16\% & Classroom       & 3.39\% & Hotel / Guestroom   & 1.70\% \\
                                &         & Library         & 1.70\% & Gym                 & 3.39\% \\
                                &         & Workshop        & 1.70\% & Lobby / Corridor    & 1.70\% \\
                                &         & Laboratory      & 1.70\% &                     &        \\
\midrule
\multicolumn{2}{c}{\textbf{Public Space}} & \multicolumn{2}{c}{\textbf{Special-Function}} & \multicolumn{2}{c}{\textbf{Outdoor Scene}} \\
\midrule
Hospital / Clinic   & 3.39\% & Parking Garage      & 3.39\% & Outdoor Environment & 10.16\% \\
School / Hall       & 3.39\% & Machinery Room      & 3.39\% &                     &         \\
Airport / Station   & 3.39\% & Elevator            & 3.39\% &                     &         \\
Museum              & 3.39\% & Corridor / Passage  & 3.39\% &                     &         \\
                    &        & Storage / Warehouse & 3.39\% &                     &         \\
\bottomrule
\end{tabular}
\vspace{-10pt}
\end{table}
To collect large-scale real-world RGB-D data, we build a scalable RGB-D capture prototype, as shown in \cref{fig:hardware-prototye}. Specifically, we design and fabricate custom mounting fixtures using 3D printing, which allows flexible attachment of different types of commercial RGB-D cameras to the rear side of the fixture. On the front side, a portable PC client equipped with a touchscreen is integrated to receive and manage the data streams from the cameras.
We develop a unified data acquisition interface using the official SDKs provided by the respective camera manufacturers. This lightweight and modular design makes the capture system both scalable and user-friendly, significantly lowering the barrier for large-scale RGB-D data collection.
We deploy multiple capture devices to scale up data acquisition over the scenes listed in \cref{tab:real_scene_distribution}. %
Because the real captures do not have missing-free depth maps, we follow FoundationStereo~\cite{FoundationStereo} to compute the stereo disparity of the left-right IR pairs to obtain the pseudo depth labels. To ensure the quality of pseudo depth maps from stereo matching, we do the left-right check and filter out the inconsistent pixel values from the depth maps.
Finally, a total of 2M real captures with extradinary scene diversity was obtained for masked depth modeling.

\subsection{Training Data Summary}

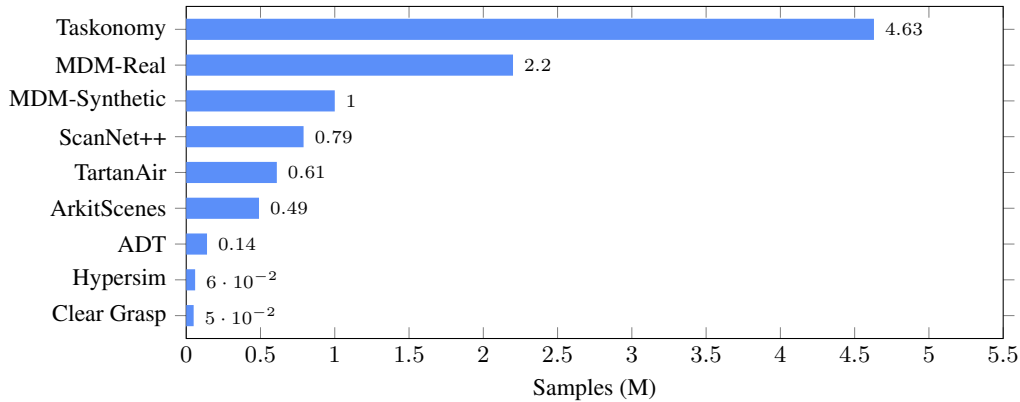
\begin{figure}[t]
    \centering
    \small
    \begin{tikzpicture}
        \begin{axis}[
            xbar,
            width=0.75\linewidth,
            height=6cm,
            bar width=8pt,
            xlabel={Samples (M)},
            xmin=0,
            xmax=5.5,
            ytick=data,
            yticklabels={Clear Grasp, Hypersim, ADT, ArkitScenes, TartanAir, ScanNet++, MDM-Synthetic, MDM-Real, Taskonomy},
            y tick label style={font=\small},
            x tick label style={font=\small},
            xlabel style={font=\small},
            nodes near coords,
            nodes near coords style={font=\scriptsize, anchor=west},
            every node near coord/.append style={xshift=1pt},
            enlarge y limits=0.08,
        ]
        \addplot[fill=pie1, draw=none] coordinates {
            (0.05,1) (0.06,2) (0.14,3) (0.49,4) (0.61,5) (0.79,6) (1.0,7) (2.2,8) (4.63,9)
        };
        \end{axis}
    \end{tikzpicture}
    \caption{\textbf{Data composition} for MDM pretraining ($\sim$10M samples total). Our self-curated data (MDM-Real and MDM-Synthetic, 3.2M) is combined with seven open-source RGB-D datasets to ensure diverse scene coverage and depth characteristics.}
    \label{tab:total-data}
\end{figure}

Besides of the self-curated 3.2M data (including the \methods and \methodr splits), we also use the following open-source datasets~\cite{clear_grasp, tartanair, hypersim, adt, scannet++, taskonomy, arkitscenes} as training data to form a total of 10M training samples for masked depth modeling. 
Note that the open-source synthetic datasets have no missing depth measurements, we randomly generate the patchified tokens to meet the expected masking ratio range of 60\%-90\% without any additional processing. For the real-world open-source datasets, the masking strategy would also be dominated by the random mask sampling because there depth maps are relative complete than our curated data.
A summary of the training data is shown in \cref{tab:total-data}.

\section{Experiments}\label{sec:exp}
In this section, we experimentally validate the effectiveness of the proposed MDM pretraining.
First, we evaluate the pretrained MDM model, \method, on the task of depth completion (DC) in \cref{sec:dc}, which is most closely aligned with our pretraining objective.
Second, we compare different pretrained backbones for monocular depth estimation by training MoGe~\cite{moge} models using either DINOv2 or our MDM-pretrained weights.
Third, we demonstrate that using our MDM-pretrained model as a foundational prior improves FoundationStereo compared to its original backbone, DepthAnythingV2~\cite{DepthAnythingV2}.
Building upon these results, in \cref{sec:application}, we showcase three downstream applications: video depth completion, online 3D point tracking, and grasp pose generation for robotic dexterous manipulation.

\subsection{Depth Completion}\label{sec:dc}
As MDM pretraining is naturally aligned with the depth completion task, we evaluate our pretrained \method model against state-of-the-art approaches, including OMNI-DC~\cite{omnidc}, PromptDA~\cite{promptda}, and PriorDA~\cite{priorda}. The evaluation checkpoints for these methods are obtained from their official repositories.
We follow two evaluation protocols, (1) Block-wise Depth Masking and (2) Sparse SfM Depth Inputs.

\paragraph{Protocol 1: Block-wise Depth Masking.}
We generate incomplete depth maps by randomly masking out spatial regions of varying sizes (blocks) from the ground-truth depth, simulating depth dropout commonly observed in consumer depth cameras. 
To further model low-quality depth measurements, we corrupt the remaining depth values with additive Gaussian noise and shot-noise-like perturbations inspired by Kinect noise models~\cite{kinect_noise}, which capture sensor-specific artifacts such as quantization effects and photon noise.
Based on the severity of masking and noise level, each dataset is divided into four difficulty levels: \emph{easy}, \emph{medium}, \emph{hard}, and \emph{extreme}. As shown in \cref{tab:ablation_depth}, our method consistently outperforms all baseline approaches across all difficulty levels, demonstrating strong robustness to both structural incompleteness and measurement noise. 
Under this protocol, three benchmark datasets, iBims~\cite{ibims}, NYUv2~\cite{silberman2012indoor}, and DIODE~\cite{diode} are used for evaluation.

\paragraph{Protocol 2: Sparse SfM Depth Inputs.} Using SfM/SLAM techniques to recover camera poses and sparse scene geometry is a common choice when the depth cameras are unavailable. Therefore, recovering the complete depth from the sparse SfM/SLAM points is valuable to many downstream applications, but yet being challenging because of the sparsity. This is a much more challenging protocol than the protocol 1.
Following OMNI-DC~\cite{omnidc}, we apply the Sparse SfM observations on the ETH-SfM dataset~\cite{eth_sfm} to evaluate the performance of our MDM-pretrained model.

\begin{figure}[!ht]
    \centering
    \includegraphics[width=1.0\linewidth]{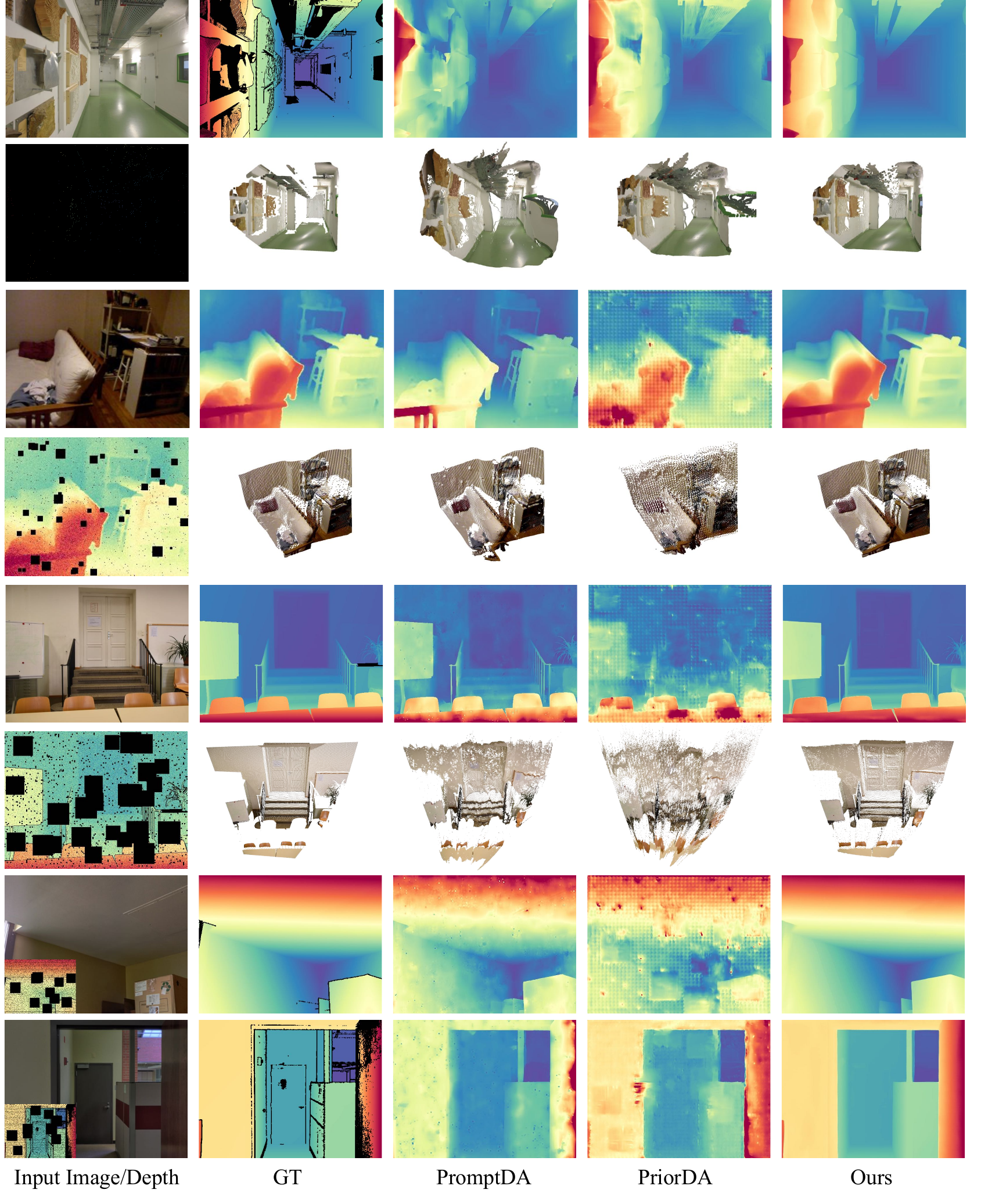}
    \vspace{-20pt}
    \caption{\textbf{Qualitative comparison of depth completion} results across four benchmarks. For each dataset, we show the RGB input, sparse/masked depth input, and predictions from OMNI-DC, PromptDA, PriorDA, and ours. Our method produces sharper boundaries and more complete structures, particularly in regions with severe occlusion or sparse observations.}
    \label{fig:dc-compare}
    \vspace{-5pt}
\end{figure}

\paragraph{Results.}
As shown in \cref{tab:ablation_depth}, our method consistently outperforms all competing approaches across all four difficulty levels on every dataset under Protocol~1.
On indoor benchmarks (iBims, NYUv2, DIODE-Indoor), \method reduces RMSE by over 40\% relative to the best competitor (PromptDA) even in the \emph{extreme} setting, demonstrating strong resilience to heavy masking and noise corruption.
On DIODE-Outdoor, where depth ranges are significantly larger, our method still achieves the lowest errors across all levels.
Under the more challenging Protocol~2 (see \cref{tab:depth_completion}), where only highly sparse SfM point clouds are available as input, \method again achieves state-of-the-art results on both indoor and outdoor splits of ETH-SfM, reducing RMSE by 47\% (indoor) and 38\% (outdoor) compared to the best baseline.
Qualitative comparisons in \cref{fig:dc-compare} further illustrate that \method produces sharper depth boundaries and more coherent structures than competing methods, particularly in regions with severe occlusion or sparse observations.
These results confirm that the MDM pretraining objective equips the model with strong depth priors that generalize across diverse corruption patterns and sparsity levels.
\begin{table}[t]
    \centering
    \caption{\textbf{Depth completion} results. (a) Block-wise depth masking with four difficulty levels on iBims, NYUv2, and DIODE. (b) Sparse SfM depth inputs on ETH3D.}
    \label{tab:depth_completion_all}
    \vspace{-2mm}

    \begin{subtable}{\linewidth}
        \centering
        \small
        \caption{Protocol 1: Block-wise Depth Masking.}
        \resizebox{\linewidth}{!}{
        \begin{tabular}{l|cc|cc|cc|cc|cc|cc|cc|cc}
        \toprule
                &  \multicolumn{2}{c|}{Easy} & \multicolumn{2}{c|}{Medium} & \multicolumn{2}{c|}{Hard} & \multicolumn{2}{c}{Extreme} &  \multicolumn{2}{c|}{Easy} & \multicolumn{2}{c|}{Medium} & \multicolumn{2}{c|}{Hard} & \multicolumn{2}{c}{Extreme} \\
                & RMSE$\downarrow$ & REL$\downarrow$ & RMSE$\downarrow$ & REL$\downarrow$ & RMSE$\downarrow$ & REL$\downarrow$ & RMSE$\downarrow$ & REL$\downarrow$ & RMSE$\downarrow$ & REL$\downarrow$ & RMSE$\downarrow$ & REL$\downarrow$ & RMSE$\downarrow$ & REL$\downarrow$ & RMSE$\downarrow$ & REL$\downarrow$\\
                \midrule
                & \multicolumn{8}{c|}{iBims} & \multicolumn{8}{c}{NYUv2}\\
                \midrule
                OMNI-DC~\cite{omnidc} & 0.476&0.084&0.929&0.232&1.382&0.342&2.053&0.555 & 0.234&0.055&0.308&0.079&0.395&0.103&0.643&0.176 \\
                OMNI-DC-DA~\cite{omnidc} & 0.602&0.068&0.771&0.155&1.266&0.302&1.937&0.538 & 0.343&0.053&0.385&0.070&0.418&0.088&0.456&0.116 \\
                PromptDA~\cite{promptda} & \underline{0.298}&\underline{0.055}&\underline{0.337}&\underline{0.065}&\underline{0.428}&\underline{0.083}&\underline{0.607}&\underline{0.129} & \underline{0.209}&\underline{0.051}&\underline{0.219}&\underline{0.053}&\underline{0.238}&\underline{0.057}&0.324&\underline{0.074} \\
                PriorDA~\cite{priorda} & 0.409&0.095&0.433&0.093&0.520&0.094&0.845&0.150 & 0.289&0.093&0.276&0.080&0.267&0.072&\underline{0.309}&0.076 \\
                Ours & \textbf{0.175}&\textbf{0.020}&\textbf{0.223}&\textbf{0.038}&\textbf{0.289}&\textbf{0.060}&\textbf{0.345}&\textbf{0.083} & \textbf{0.089}&\textbf{0.014}&\textbf{0.124}&\textbf{0.021}&\textbf{0.146}&\textbf{0.027}&\textbf{0.181}&\textbf{0.033} \\
                \midrule
                & \multicolumn{8}{c|}{DIODE-Indoor} & \multicolumn{8}{c}{DIODE-Outdoor}\\
                \midrule
                OMNI-DC~\cite{omnidc} & 0.563&0.074&0.895&0.121&1.108&0.153&1.775&0.271 & \underline{2.214}&\textbf{0.041}&2.806&\underline{0.067}&3.701&0.103&6.239&0.204 \\
                OMNI-DC-DA~\cite{omnidc} & 0.710&0.065&0.991&0.097&1.222&0.121&1.592&0.211 & 2.229&0.053&3.703&0.070&5.352&0.110&5.988&0.173 \\
                PromptDA~\cite{promptda} & \underline{0.250}&\underline{0.049}&\underline{0.291}&\underline{0.054}&\underline{0.320}&\underline{0.060}&\underline{0.465}&\underline{0.083} & 2.587&0.070&2.807&0.087&3.026&0.100&\underline{4.313}&0.156 \\
                PriorDA~\cite{priorda} & 0.370&0.073&0.354&0.064&0.369&0.062&0.665&\underline{0.083} & 2.734&0.090&\underline{2.693}&0.085&\underline{2.909}&\underline{0.088}&5.114&\underline{0.136} \\
                Ours & \textbf{0.090}&\textbf{0.013}&\textbf{0.185}&\textbf{0.028}&\textbf{0.195}&\textbf{0.031}&\textbf{0.221}&\textbf{0.032} & \textbf{2.011}&\underline{0.049}&\textbf{2.338}&\textbf{0.061}&\textbf{2.578}&\textbf{0.066}&\textbf{3.811}&\textbf{0.092} \\
        \bottomrule
        \end{tabular}
        }
        \label{tab:ablation_depth}
    \end{subtable}

    \vspace{3mm}

    \begin{subtable}{\linewidth}
        \centering
        \small
        \caption{Protocol 2: Sparse SfM Depth Inputs (ETH3D).}
        \begin{tabular}{l|cccc|cccc}
        \toprule
                &  \multicolumn{4}{c|}{Indoor}& \multicolumn{4}{c}{Outdoor}\\
                & RMSE$\downarrow$ & MAE$\downarrow$ & REL$\downarrow$ & $\delta_1\uparrow$& RMSE$\downarrow$ & MAE$\downarrow$ &REL$\downarrow$ & $\delta_1\uparrow$\\
        \midrule
        OMNI-DC~\cite{omnidc} & 0.605&0.239&0.090&0.932 & \underline{1.069}&0.312&0.053&0.954 \\
        OMNI-DC-DA~\cite{omnidc} & 0.489&0.164&0.061&\underline{0.967} & 1.093&\textbf{0.215}&\underline{0.035}&\textbf{0.979} \\
        PromptDA~\cite{promptda} & 1.023&0.648&0.242&0.796 & 2.750&1.612&0.300&0.776 \\
        PriorDA~\cite{priorda} & \underline{0.360}&\underline{0.157}&\underline{0.058}&0.963 & 1.238&0.459&0.090&0.952 \\
        \midrule
        Ours & \textbf{0.192}&\textbf{0.081}&\textbf{0.023}&\textbf{0.982} & \textbf{0.664}&\underline{0.223}&\textbf{0.033}&\underline{0.976} \\
        \bottomrule
        \end{tabular}
        \label{tab:depth_completion}
    \end{subtable}
\end{table}

\subsection{Monocular Depth Estimation}
Monocular depth estimation has benefited significantly from large-scale visual pretraining~\cite{DINOv2}. As a new pretraining paradigm that learns joint representations of RGB appearance and depth measurements, we use our \method model as an alternative to DINOv2~\cite{DINOv2} for initializing MoGe~\cite{moge}. To adapt our pretrained model for monocular input, we remove the depth embedding branch and the ConvStack decoder, retaining only the encoder backbone for RGB images.

\paragraph{Training.}
We follow the standard fine-tuning protocols, and train two MoGe models (one with DINOv2 pretrain, another one with \method pretrain). To save the training time, we only use the TartanAir dataset~\cite{tartanair} in the training instead of the official implementation of MoGe~\cite{moge} that use much more datasets to boost the performance. 

\paragraph{Evaluation Protocol.} The trained models are then evaluated on a wide range of diverse benchmarks, including indoor (NYUv2~\cite{silberman2012indoor}, HAMMER~\cite{hammer}), outdoor driving (KITTI~\cite{kitti}, DDAD~\cite{ddad}), large-scale synthetic (GSO~\cite{gso}, Sintel~\cite{sintel}), and challenging real-world scenes (iBims-1~\cite{ibims}, DIODE~\cite{diode}, Spring~\cite{spring}).

\paragraph{Results.}
As shown in \cref{tab:monocular_depth}, models initialized with our encoder achieve consistent and significant improvements across all datasets compared to those based on DINOv2, demonstrating stronger generalization and enhanced intrinsic spatial understanding. By learning robust spatial representations through masked depth modeling, the encoder enables accurate depth reasoning even without depth inputs at inference time. These results confirm that our pretraining paradigm effectively distills 3D geometric knowledge into the encoder, improving its ability to infer depth structure from monocular images.

\begin{table}[t]
        \centering
        \caption{\textbf{Monocular depth estimation} results using MoGe with different pretrained backbones (DINOv2 vs.\ ours). We evaluate depth and point map accuracy under affine-invariant, scale-invariant, and disparity-invariant metrics across 10 diverse benchmarks.}
        \resizebox{\linewidth}{!}{
        \begin{tabular}{l|cc|cc|cc|cc|cc|cc|cc|cc|cc|cc}
        \toprule
            & \multicolumn{6}{c|}{Depth} & \multicolumn{4}{c|}{Point} & \multicolumn{6}{c|}{Depth} & \multicolumn{4}{c}{Point}\\
            & \multicolumn{2}{c|}{Aff-inv} & \multicolumn{2}{c|}{Scl-inv} & \multicolumn{2}{c|}{Dsp-inv} & \multicolumn{2}{c|}{Aff-inv} & \multicolumn{2}{c|}{Scl-inv} & \multicolumn{2}{c|}{Aff-inv} & \multicolumn{2}{c|}{Scl-inv} & \multicolumn{2}{c|}{Dsp-inv} & \multicolumn{2}{c|}{Aff-inv} & \multicolumn{2}{c}{Scl-inv}\\
            & REL$\downarrow$ & $\delta_1\uparrow$ & REL$\downarrow$ & $\delta_1\uparrow$ & REL$\downarrow$ & $\delta_1\uparrow$ & REL$\downarrow$ & $\delta_1\uparrow$ & REL$\downarrow$ & $\delta_1\uparrow$ & REL$\downarrow$ & $\delta_1\uparrow$ & REL$\downarrow$ & $\delta_1\uparrow$ & REL$\downarrow$ & $\delta_1\uparrow$ & REL$\downarrow$ & $\delta_1\uparrow$ & REL$\downarrow$ & $\delta_1\uparrow$ \\
            \midrule
            \midrule
            & \multicolumn{10}{c|}{NYUv2} & \multicolumn{10}{c}{KITTI} \\
            \midrule
    DINOv2 & 0.056&0.965&0.068&0.948&0.061&0.966&0.066&0.957&0.068&0.948 & 0.071&0.934&0.226&0.544&0.154&0.835&\textbf{0.204}&\textbf{0.615}&0.228&0.539 \\
    Ours & \textbf{0.044}&\textbf{0.976}&\textbf{0.054}&\textbf{0.966}&\textbf{0.049}&\textbf{0.978}&\textbf{0.052}&\textbf{0.971}&\textbf{0.054}&\textbf{0.966} & \textbf{0.066}&\textbf{0.936}&\textbf{0.220}&\textbf{0.556}&\textbf{0.144}&\textbf{0.859}&0.208&0.599&\textbf{0.222}&\textbf{0.549} \\
            \midrule
            & \multicolumn{10}{c|}{ETH3D} & \multicolumn{10}{c}{iBims-1} \\
            \midrule
    DINOv2 & 0.064&0.948&0.137&0.790&0.152&0.780&0.127&0.837&0.137&0.789 & 0.054&0.963&0.112&0.856&0.071&0.947&0.104&0.874&0.112&0.854 \\
    Ours & \textbf{0.053}&\textbf{0.963}&\textbf{0.091}&\textbf{0.902}&\textbf{0.075}&\textbf{0.939}&\textbf{0.088}&\textbf{0.923}&\textbf{0.092}&\textbf{0.900} & \textbf{0.044}&\textbf{0.973}&\textbf{0.091}&\textbf{0.899}&\textbf{0.060}&\textbf{0.962}&\textbf{0.085}&\textbf{0.924}&\textbf{0.091}&\textbf{0.899} \\
            \midrule
            & \multicolumn{10}{c|}{GSO} & \multicolumn{10}{c}{Sintel} \\
            \midrule
    DINOv2 & 0.024&0.999&0.028&0.998&0.024&0.999&0.028&0.998&0.028&0.997 & 0.240&0.664&0.318&0.554&0.425&0.529&0.310&0.559&0.318&0.554 \\
    Ours & \textbf{0.018}&\textbf{1.000}&\textbf{0.026}&\textbf{1.000}&\textbf{0.019}&\textbf{1.000}&\textbf{0.026}&\textbf{1.000}&\textbf{0.026}&\textbf{1.000} & \textbf{0.204}&\textbf{0.715}&\textbf{0.305}&\textbf{0.580}&\textbf{0.342}&\textbf{0.604}&\textbf{0.298}&\textbf{0.573}&\textbf{0.305}&\textbf{0.580} \\
            \midrule
            & \multicolumn{10}{c|}{DDAD} & \multicolumn{10}{c}{DIODE} \\
            \midrule
    DINOv2 & 0.147&0.789&0.230&\textbf{0.598}&0.224&0.703&\textbf{0.215}&\textbf{0.647}&0.231&\textbf{0.595} & 0.071&0.927&0.179&0.731&0.128&0.849&0.159&\textbf{0.774}&0.179&0.730 \\
    Ours & \textbf{0.135}&\textbf{0.810}&0.230&0.594&\textbf{0.212}&\textbf{0.742}&0.219&0.626&0.231&0.589 & \textbf{0.061}&\textbf{0.942}&\textbf{0.169}&\textbf{0.751}&\textbf{0.105}&\textbf{0.893}&0.159&0.762&\textbf{0.169}&\textbf{0.751} \\
            \midrule
            & \multicolumn{10}{c|}{Spring} & \multicolumn{10}{c}{HAMMER} \\
            \midrule
    DINOv2 & 0.277&0.606&\textbf{0.380}&0.434&0.472&0.415&\textbf{0.364}&\textbf{0.457}&\textbf{0.380}&0.434 & 0.061&0.954&0.091&0.915&0.068&0.954&0.089&0.924&0.091&0.914 \\
    Ours & \textbf{0.245}&\textbf{0.638}&0.385&\textbf{0.436}&\textbf{0.431}&\textbf{0.495}&0.373&0.430&0.385&\textbf{0.435} & \textbf{0.053}&\textbf{0.962}&\textbf{0.080}&\textbf{0.933}&\textbf{0.060}&\textbf{0.967}&\textbf{0.079}&\textbf{0.941}&\textbf{0.080}&\textbf{0.932} \\
        \bottomrule
        \end{tabular}
        }
        \label{tab:monocular_depth}
    \end{table}

\begin{figure}[t]
    \centering
    \includegraphics[width=\linewidth]{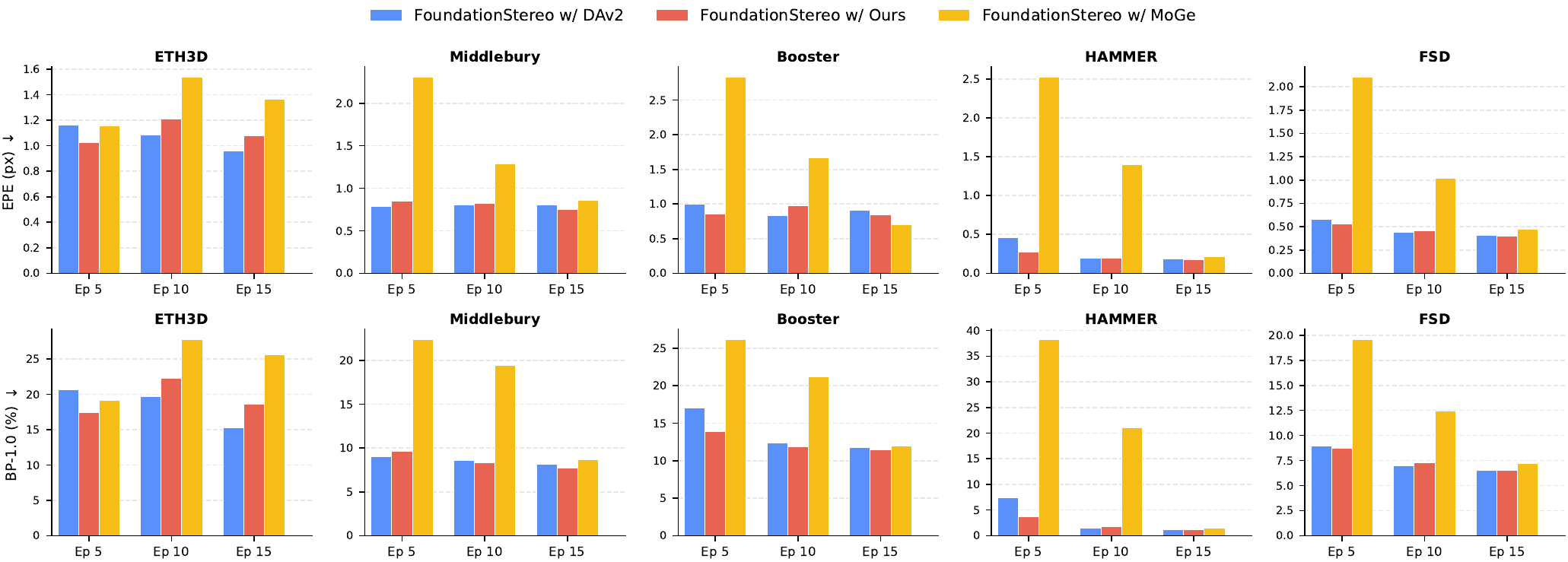}
    \caption{\textbf{Epoch-wise comparison of FoundationStereo} trained with different depth priors. We report EPE (top) and BP-1.0 (bottom) at epochs 5, 10, and 15 across five benchmarks.}
    \label{fig:fstereo-train}
\end{figure}

\subsection{FoundationStereo with MDM Pretraining}
We also use our MDM-pretrained \method model as a strong monocular depth prior in FoundationStereo~\cite{FoundationStereo}. Following the meta architecture, we adapt the RGB-only branch of \method model as the foundational depth priors and keep the rest of the architectures the same as their official implementation. Because our model architecture in MDM is similar to MoGe~\cite{moge2}, we also use MoGe's official implementation as an altenrative to train a FoundationStereo with DepthAnythingV2 for comparison. Besides, we also train the vanilla FoundationStereo to obtain epoch-wise comparisons. 
The FSD dataset~\cite{FoundationStereo} is used for training. We keep all training hyperparameters the same for these three runs to make comparisons. To save the compute resource of training, we do not do iterative self data curation as what it was done in their original implementation. The total epochs of each run is 15.

\paragraph{Evaluation Protocol.} We follow the evaluation protocol of the three trained models on ETH3D~\cite{eth_sfm}, Middlebury~\cite{middlebury}, Booster~\cite{booster}, Hammer~\cite{hammer} and FoundationStereo dataset~\cite{FoundationStereo} and report the EPE (end-point-error ) and BP-1.0 that computes the percentage of pixels where the
disparity error is larger than 1.0 pixel. 

\paragraph{Results.} We report the performance of each model at 5-th, 10-th and 15-th epochs in \cref{fig:fstereo-train}. Several key observations emerge: (1) \textbf{Faster convergence.} At epoch 5, FoundationStereo with our pretrained encoder already achieves strong performance, outperforming the vanilla baseline on most benchmarks (\emph{e.g.}, HAMMER EPE: 0.27 vs.\ 0.46, Booster EPE: 0.86 vs.\ 1.00). (2) \textbf{Training stability.} The MoGe-based variant exhibits significant instability in early training, with dramatically higher errors at epoch 5 (\emph{e.g.}, HAMMER EPE: 2.53, Booster EPE: 2.84) that persist through epoch 10. (3) \textbf{Final performance.} At epoch 15, our variant achieves the best or comparable results across all benchmarks (Middlebury EPE: 0.75, HAMMER EPE: 0.17, FSD EPE: 0.40), confirming that MDM pretraining provides a more effective initialization for downstream stereo matching.

\section{Extensions and Applications}\label{sec:application}
In this section, we mainly showcase the extensions and applications built atop of our \method with a Orbbec Gemini-335 camera as the RGB-D input. 
There is no post-training in any of the experiments in this section.
We hope our preliminary extensions could answer a question of why are RGB-D cameras important and how will RGB-D cameras shape the future in spatial perception.

\subsection{Video Depth Completion}
Even though our MDM pretraining was conducted on image data instead of the video inputs, we found that our \method model could significantly reduce the spatial-temporal inconsistency of the RGB-D cameras in several challenging scenarios.
\begin{figure}[t]
    \centering
    \includegraphics[width=\linewidth]{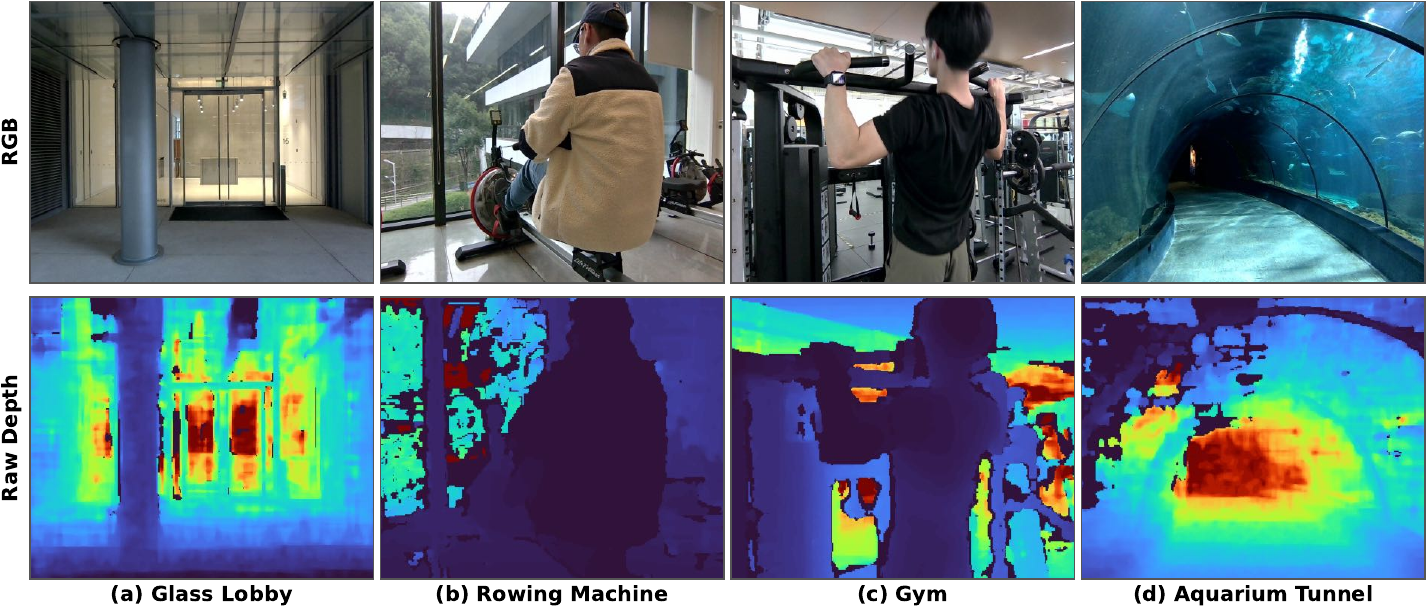}
    \caption{\textbf{Video depth completion results under challenging scenarios.} The raw depth maps from the Orbbec sensor exhibit significant missing regions (shown in black) around transparent and reflective surfaces such as glass walls, windows, mirrors, and aquarium tunnels.}
    \label{fig:video-depth-input}
\end{figure}
\Cref{fig:video-depth-input} shows four challenging scenarios ({\em i.e.,} glass lobby, rowing machine, gym and aquarium tunnels) that are captured by ourselves. These videos are captured at 30 FPS in 640$\times 480$ resolution.

\begin{figure}[t]
    \centering
    \includegraphics[width=\linewidth]{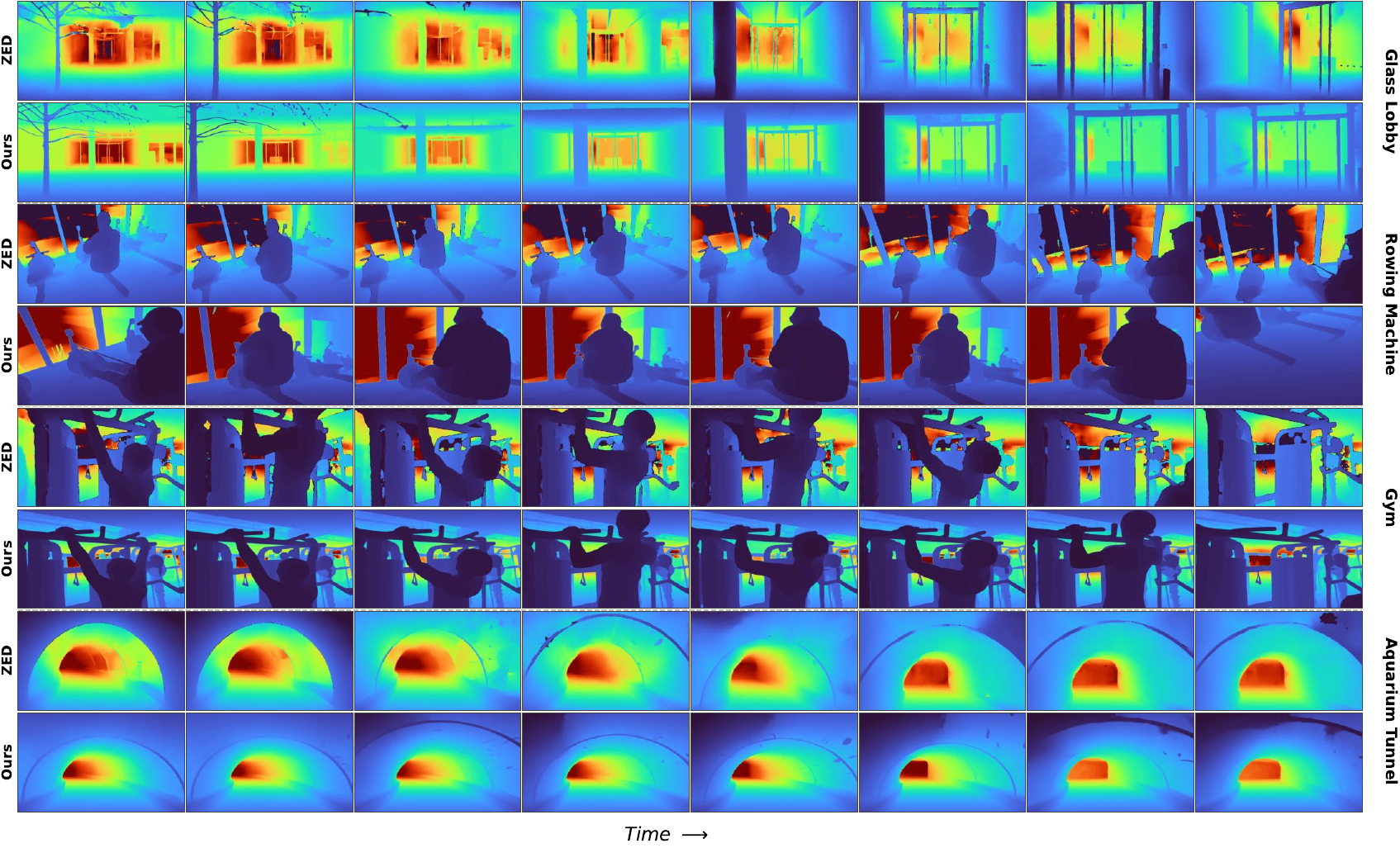}
    \vspace{-20pt}
    \caption{\textbf{Video depth completion} results compared with ZED stereo depth. For each scenario, the top row shows ZED depth and the bottom row shows our predictions. The ZED sensor suffers from significant missing regions on transparent and reflective surfaces, while our model produces spatially complete and temporally consistent depth maps throughout all sequences.}
    \label{fig:video-depth-strip}
    \vspace{-10pt}
\end{figure}

We make comparisons to the ZED-mini camera that is co-mounted in our portable capture system. 
As shown in \cref{fig:video-depth-strip}, we compare our \method predictions against the ZED stereo camera depth, which serves as a higher-quality reference. Our model significantly improves upon the raw Orbbec depth by: (1) \emph{filling large missing regions} caused by transparent and reflective surfaces where both structured-light (Orbbec) and stereo (ZED) sensors fail, such as glass walls in the lobby, windows near the rowing machine, mirrors in the gym, and the aquarium tunnel; (2) \emph{recovering fine-grained structures} including thin objects (e.g., exercise equipment bars, ceiling pipes) and sharp object boundaries that are often noisy or absent in raw sensor outputs; and (3) \emph{maintaining temporal consistency} across frames without explicit temporal modeling or video-specific training, producing smooth and stable depth sequences despite the per-frame inference. Notably, in the aquarium tunnel scenario, the ZED stereo camera almost entirely fails due to the refractive glass surfaces, while our model produces geometrically plausible depth throughout the sequence.

\subsection{Online 3D Point Tracking}
With the excellent performance on video depth completion, we leverage an off-the-shelf SpatialTrackerV2~\cite{spatracker2} to obtain the camera poses and optionally track 3D points on dynamic objects. In our implementation, we use the online version of SpatialTrackerV2~\cite{spatracker2} that does not depend on the VGGT~\cite{vggt}-Frontend for initial pose and depth estimation as the base, and tailored it as an RGB-D tracking baseline. Different from the SpatialTrackerV2's original implementation that assumed the RGB-D inputs are with known camera extrinsics, we initialize the framewise extrinsics using SE-3 identity and mainly resort to the Bundle Adjustment (BA) on the tracked VO points to track the camera motion. There is no global BA applied to enable the efficiency. Besides, we did not finetune the SpatialTrackerV2.

\begin{figure}[!ht]
    \centering
    \begin{subfigure}[t]{\linewidth}
        \centering
        \includegraphics[width=\linewidth]{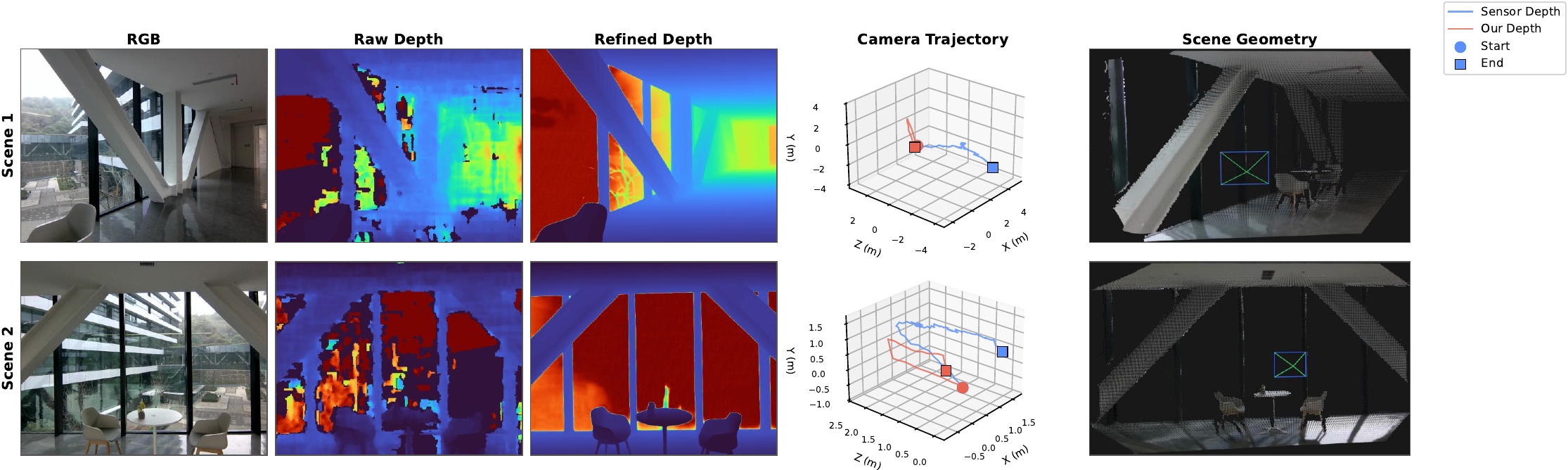}
        \caption{Camera tracking and scene reconstruction. From left to right: RGB input, raw sensor depth, our refined depth, estimated camera trajectories (sensor depth in blue vs.\ our depth in red), and the reconstructed scene geometry.}
        \label{fig:scene-tracking}
    \end{subfigure}
    \\[6pt]
    \begin{subfigure}[t]{\linewidth}
        \centering
        \includegraphics[width=\linewidth]{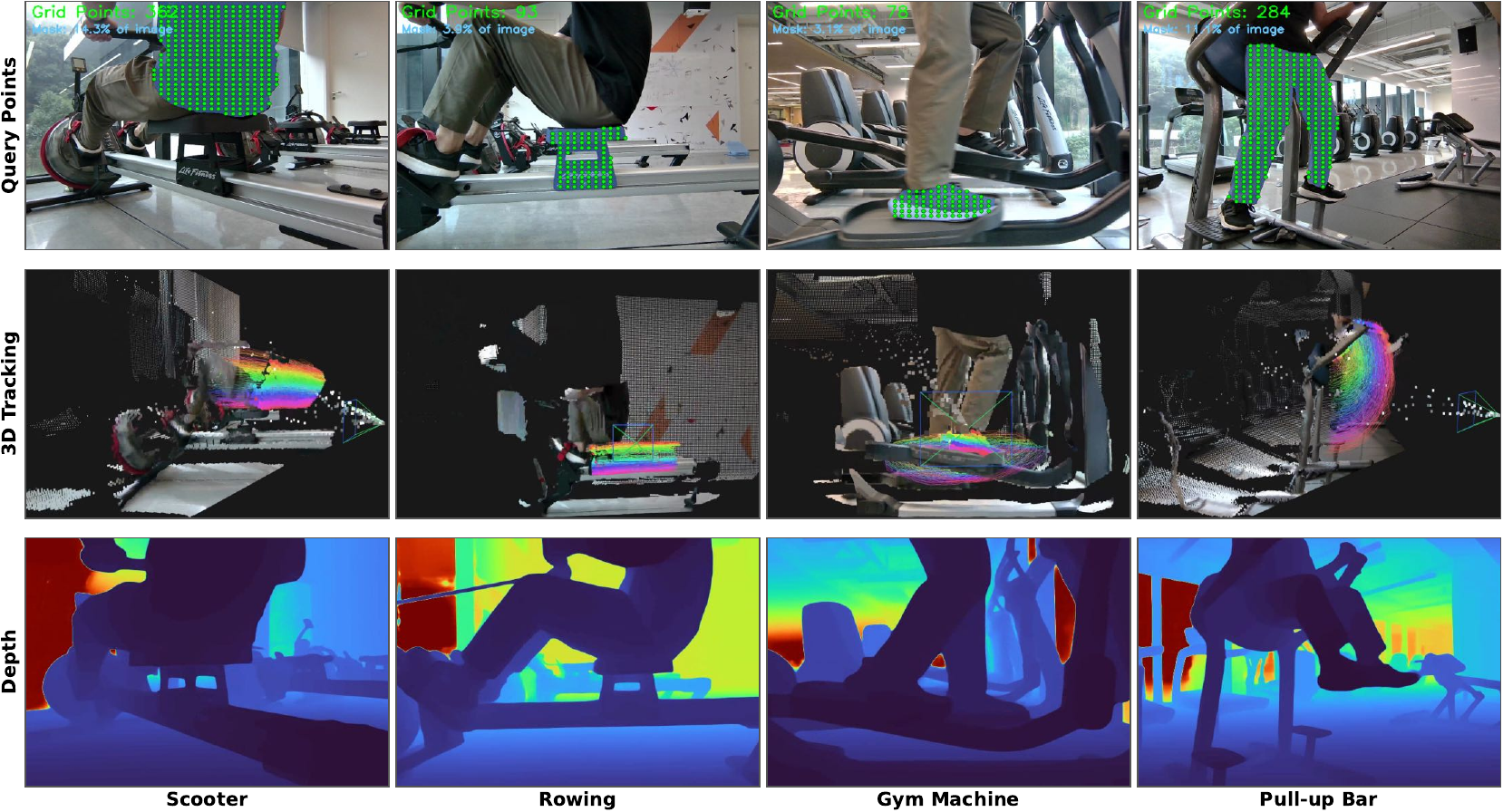}
        \caption{Dynamic 3D point tracking. Top: query points on the target object. Middle: 3D tracked trajectories (rainbow-colored by time). Bottom: corresponding depth maps.}
        \label{fig:dynamic-tracking}
    \end{subfigure}
    \caption{\textbf{Online 3D point tracking} results using SpatialTrackerV2 with our refined depth.}
    \label{fig:3d-tracking}
\end{figure}

\paragraph{Camera Tracking.}
As shown in \cref{fig:scene-tracking}, we evaluate camera tracking on two indoor scenes with extensive glass surfaces where the raw depth sensor fails severely. Using the refined depth from our model, the SpatialTrackerV2 produces significantly smoother and more accurate camera trajectories compared to using raw sensor depth, which suffers from severe drift due to the missing depth regions.

\paragraph{Object Motion Tracking.}
We further demonstrate dynamic 3D point tracking on four scenarios with moving objects, as shown in \cref{fig:dynamic-tracking}. Given a set of query points on the object of interest, SpatialTrackerV2 tracks their 3D trajectories using our refined depth. The rainbow-colored trails in the 3D visualizations reveal coherent motion patterns for each object, confirming that our depth predictions maintain sufficient geometric accuracy for downstream dynamic tracking tasks.

\subsection{Grasp Pose Generation in Real-World Robotics}
We apply our \method to a real-world dexterous grasping pipeline, where accurate depth is critical for generating precise grasp poses.

\begin{figure}[t]
    \centering
        \includegraphics[height=.19\textheight]{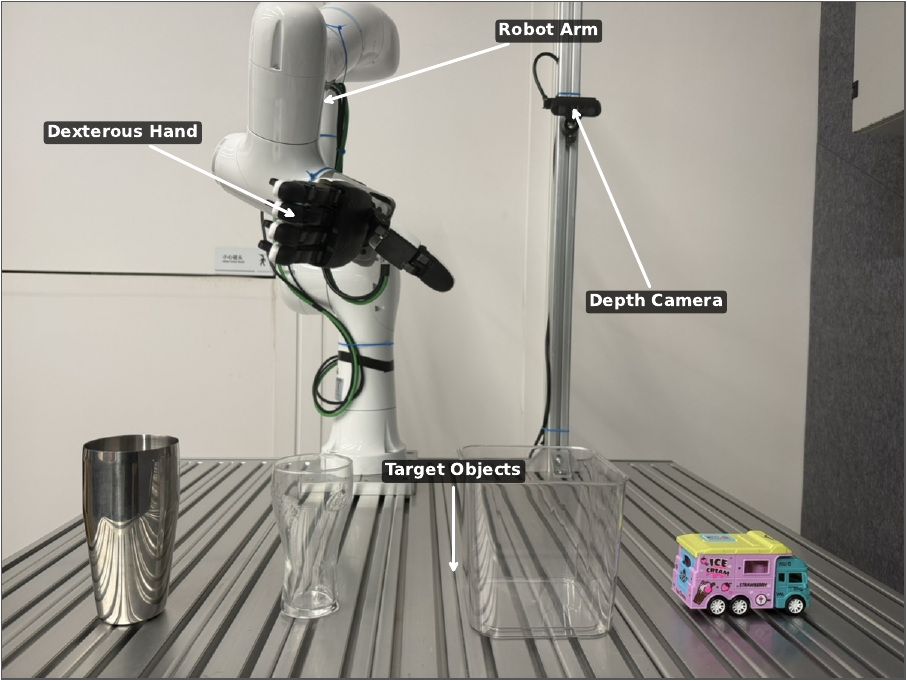}
        \includegraphics[height=.19\textheight]{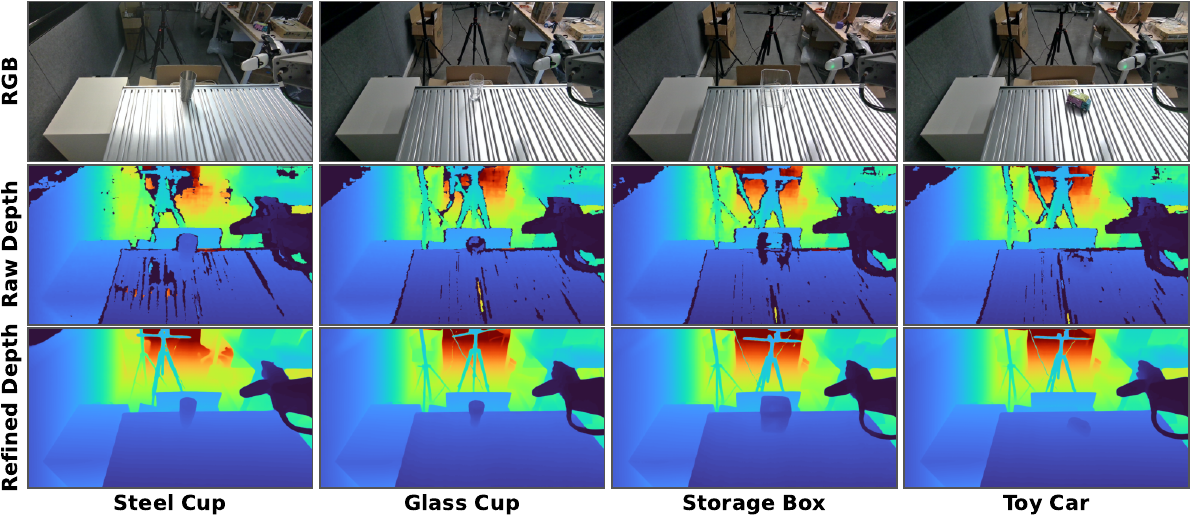}
    \caption{\textbf{Qualitative results of the grasping experiment.} Left: hardware setup with the robotic arm, dexterous gripper, and depth camera. Right: RGB, raw sensor depth, and our refined depth for four target objects. The raw depth is severely corrupted for reflective (steel cup) and transparent (glass cup, storage box) objects, while our method produces complete, geometrically accurate depth maps.}
    \label{fig:grasp}
    \vspace{-5pt}
\end{figure}
\paragraph{Setup.}
Our system consists of a Rokae XMate-SR5 robotic arm equipped with an X~Hand-1 dexterous hand, and an Orbbec Gemini~335 RGB-D camera for perception. Given an RGB-D observation, we first convert the depth into a point cloud and then predict an $N\times22$ dexterous hand pose using a diffusion policy. The policy is conditioned on RGB features extracted by DINOv2 (ViT-L/14) and point cloud features from a Point Transformer, following a DP3-like architecture~\cite{dp3}. The model is trained on the HOI4D dataset~\cite{hoi4d}, where human hand-object interactions are retargeted to the dexterous hand configuration via 3D keypoint correspondence.

\paragraph{Results.}
\begin{figure}[t]
    \centering
    \includegraphics[width=\linewidth]{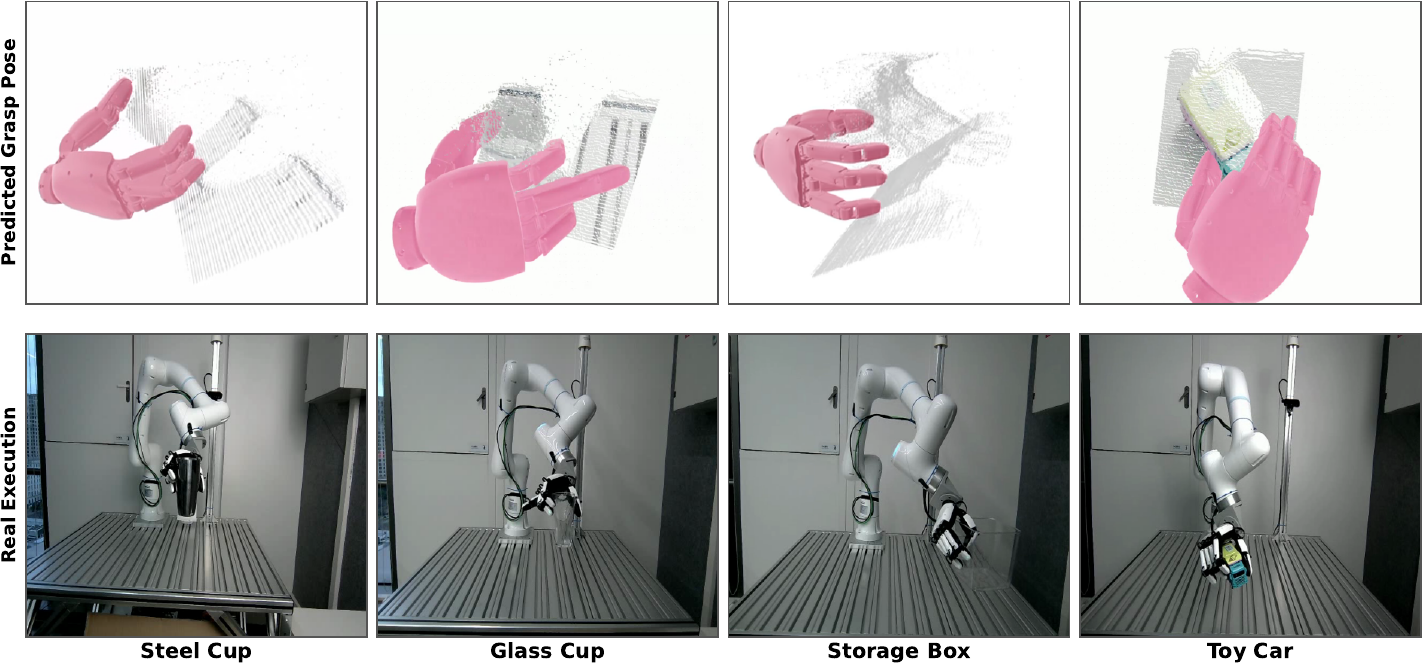}
    \caption{\textbf{Grasp pose generation and real-world execution.} Top: predicted grasp poses rendered as a dexterous hand overlaid on the point cloud reconstructed from our refined depth. Bottom: successful grasps executed by the robotic system on each target object.}
    \label{fig:grasp-demo}
\end{figure}

\begin{table}[!ht]
    \centering
    \caption{\textbf{Grasping success rates} (out of 20 trials) using our refined depth vs.\ raw sensor depth on four challenging objects. The transparent storage box is entirely ungraspable with raw depth due to severe depth corruption.}
    \vspace{-5pt}
    \setlength{\tabcolsep}{18pt}
    \small
    \begin{tabular}{l|ccc}
    \toprule
        &  Total Times & Grasp with \method & Grasp with raw depth\\
    \midrule
    Stainless steel cup & 20 & 17 & 13\\
    Transparent cup & 20 & 16 & 12 \\
    Toy Car & 20 & 16 & 9 \\
    Transparent storage box & 20 & 10 & N/A  \\
    \bottomrule
    \end{tabular}
    \label{tab:grasping}
    \vspace{-10pt}
\end{table}
We evaluate grasping performance on four challenging objects, including transparent and reflective items for which raw depth sensors typically fail. \Cref{fig:grasp-demo} shows the predicted grasp poses and their interaction with the point clouds reconstructed from our completed depth maps.
As reported in \cref{tab:grasping}, \method consistently improves grasping success rates compared to using raw sensor depth. Notably, the transparent storage box is entirely ungraspable with raw depth (N/A) due to severe depth corruption, whereas our model achieves a 50\% success rate by producing geometrically plausible depth estimates, despite occasional inaccuracies on highly transparent surfaces. These results demonstrate that improved depth completion directly translates into more reliable robotic manipulation in real-world scenarios.

\section{Conclusion}\label{sec:conclusion}
In this technical report, we investigate the problem of missing depth measurements in RGB-D cameras. Rather than treating these missing values purely as sensor failures, we leverage them as natural masks that reflect appearance ambiguities inherent in depth imaging systems, and propose an effective solution termed Masked Depth Modeling (MDM). Using 3 million self-curated RGB-D samples together with open-source depth datasets, we pretrain a large Vision Transformer to learn joint representations of pixel appearance and depth measurements. The resulting model, \method, demonstrates strong performance on both depth completion and monocular depth estimation, and further serves as a powerful monocular depth prior for FoundationStereo.
Beyond static image tasks, we show that \method model produces high-quality video depth streams with strong spatial and temporal consistency. We further validate its effectiveness in downstream applications, including 3D point tracking and dexterous robotic grasping.

\section*{Acknowledgments}
We gratefully acknowledge the optical testing team at Orbbec Inc. for their professional support.

{
\small
\bibliographystyle{plain}
\bibliography{ref.bib}
}

\end{document}